\documentclass[final]{template}

\usepackage{times}
\usepackage{epsfig}
\usepackage{graphicx}
\usepackage{amsmath}
\usepackage{amssymb}
\usepackage{mathtools}
\usepackage{url}
\usepackage{color}
\usepackage{float}
\usepackage{array}
\usepackage{multirow}
\usepackage{arydshln}
\usepackage{longtable}
\usepackage{booktabs}

\usepackage{lipsum}
\usepackage{adjustbox}
\usepackage{enumitem}
\usepackage[pagebackref=true,breaklinks=true,colorlinks,bookmarks=false]{hyperref}

\begin{document}

\title{Fast Neural Architecture Search for Lightweight Dense Prediction Networks}

\author{Lam Huynh$^1$ \qquad
Esa Rahtu$^2$ \qquad
Jiri Matas$^3$ \qquad
Janne Heikkil\"a$^1$ \\
\small{$^1$Center for Machine Vision and Signal Analysis, University of Oulu} \\ 
\small{$^2$Computer Vision Group, Tampere University} \\
\small{$^3$ Center for Machine Perception, Czech Technical University, Czech Republic} }

\maketitle

\begin{abstract}
We present LDP, a lightweight dense prediction neural architecture search (NAS) framework. Starting from a pre-defined generic backbone, LDP applies  the novel Assisted Tabu Search for efficient architecture exploration. LDP is fast and suitable for various dense estimation problems, unlike previous NAS methods that are either computational demanding or deployed only for a single subtask. The performance of LPD is evaluated on monocular depth estimation, semantic segmentation, and image super-resolution tasks on diverse datasets, including NYU-Depth-v2, KITTI, Cityscapes, COCO-stuff, DIV2K, Set5, Set14, BSD100, Urban100. Experiments show that the proposed framework yields consistent improvements on all tested dense prediction tasks, while being $5\%-315\%$ more compact in terms of the number of model parameters than prior arts.
\end{abstract}

Dense prediction is a class of computer vision problems aiming at mapping every pixel of the input image with some predicted values. Depending on the problem, the output values can be either continous or discrete. For instance, monocular depth estimation and image super-resolution are often formulated as regression, while semantic segmentation is a dense classification, i.e. discrete,  problem.
More specifically, the monocular depth estimation problem produces a dense depth map from a single image to be used in various applications including robotics, scene understanding, and augmented reality. Single image super-resolution (SISR) is a low-level vision task that generates a high-resolution image from its low-resolution counterpart. SISR is widely utilized in medical and surveillance imaging, where images with more precise details can provide invaluable information. On the other hand, semantic segmentation predicts a dense annotated map of different semantic categories from a given image that is crucial for image understanding tasks.

Recent deep neural networks (DNN) exhibits remarkable results on dense prediction, especially subproblems such as single image depth estimation~\cite{bhat2021adabins,chen2019structure,facil2019cam,Hu2018RevisitingSI,huynh2020guiding,lee2019big,liu2019planercnn,liu2018planenet,ramamonjisoa2019sharpnet,yang2021transformers,chen2016single,lee2019monocular,Ranftl2021,garg2016unsupervised,godard2019digging}, semantic segmentation~\cite{howard2019searching,chen2019fasterseg,li2019dfanet,orsic2019defense,wang2020deep,yu2021lite}, and single image super-resolution~\cite{dong2016accelerating,behjati2021overnet,tai2017memnet}.

Nonetheless, state-of-the-art methods mainly enhance dense prediction accuracy by increasing network complexity hindering the applicability on resource limited devices. Vision transformer-based approaches~\cite{Ranftl2021,yang2021transformers,liu2021swin,jain2021semask} achieve state-of-the-art results but are subservient to large model and massive data from training.

The most straightforward way to deal with the computational limitations is to use simple and small architectures~\cite{wofk2019fastdepth}. Usually such simple designs are unreliable and yield low-quality predictions. Other popular strategies include quantizing the weights of a network into low-precision fixed-point operations~\cite{han2015deep} or pruning by directly cutting off less important filters~\cite{yang2018netadapt}. Nevertheless, these methods depend on a baseline model, tend to degrade their performance afterwards and are incapable of exploring new combinations of DNN operations. Moreover, creating a resource-constrained model is a non-trivial task requiring 1) expert knowledge to carefully balance accuracy and resources and 2) plenty of tedious trial-and-error work.

\begin{figure}[!t]
\begin{center}

  \includegraphics[width=0.84\linewidth]{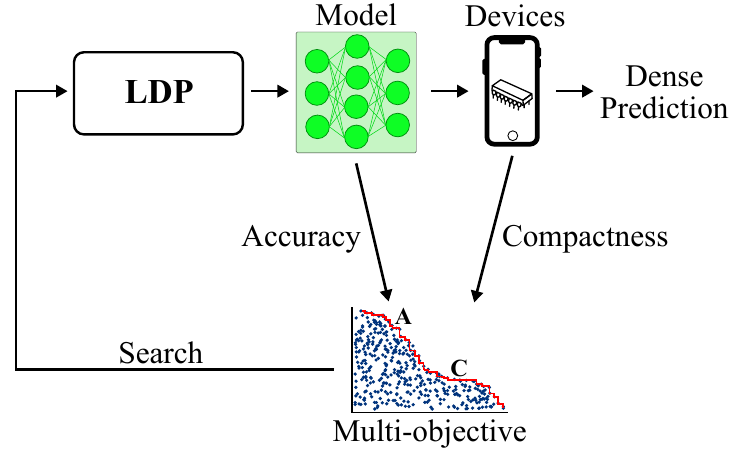}
\end{center}
  \vspace{-0.35cm}
  \caption{The proposed framework enables efficient NAS for various lightweight dense prediction tasks including depth prediction, semantic segmentation and image super-resolution.}
  \vspace{-0.35cm}
\label{fig:overview}
\end{figure}

\begin{figure*}[!t]
\begin{center}
  \includegraphics[width=0.99\linewidth]{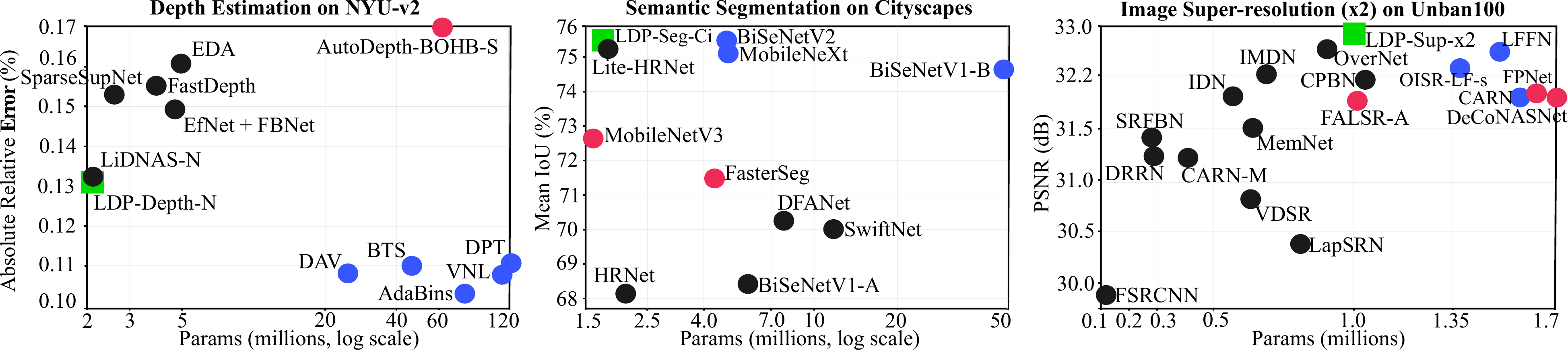}
\end{center}
\vspace{-0.35cm}
  \caption{
 From left to right, absolute relative error, mean intersection-over-union and peak signal-to-noise ratio vs.~the number of parameters for recent dense prediction methods on NYU-Depth-v2 (left), Cityscapes (middle) and Urban100 (right) -- the LDP models outperforms the lightweight baselines (black), while using substantially less parameters than the current state-of-the-art methods (blue). Compared to the recent NAS-based approaches (red), LDP improves in both performance and compactness.
  } \vspace{-0.35cm}
\label{fig:rel_params_chartv1} %
\end{figure*}

Neural architecture search (NAS), proposed recently~\cite{zoph2016neural,zoph2018learning}, exhibits compelling results, and more importantly, promises to release from manual tweaking of deep neural architectures. Unfortunately, NAS methods mostly require thousands of training hours on hundreds of GPUs. To address this, recent NAS studies introduced various techniques to increase the efficiency, including weight sharing~\cite{pham2018efficient}, and network transformation~\cite{elsken2018efficient}. These methods show promising results, but they are still expensive and mainly focus on classification and detection. 

This paper introduces LDP, which is a novel multi-objective NAS framework, capable of searching for accurate and lightweight dense prediction architectures as shown in Figure~\ref{fig:overview}. 
The approach is based on three core ideas. First, the previous NAS methods essentially search for a few types of cells and then repeatedly accumulate the same cells to build the whole network. Although doing this simplifies the search process, it also restrains layer diversity that is important for computational efficiency. On the other hand, multi-scale pyramid network structure has been successful for many dense prediction tasks including optical flow~\cite{sun2018pwc}, depth estimation~\cite{poggi2018towards}, semantic segmentation~\cite{zhao2017pyramid}, and image super-resolution~\cite{ahn2018fast}. Based on these observations, we construct a generic pre-defined backbone network that utilizes different layers striving for the right balance between flexibility and search space size.

Second, we propose the Assisted Tabu Search (ATS) for efficient neural architecture search. Inspired by the recent NAS study that suggests estimating network performance without training~\cite{mellor2021neural}, we integrate this idea into our multi-objective search function to swiftly evaluate our candidate networks. This, in turn, significantly reduces search time compared to state-of-the-art NAS-based approaches~\cite{saikia2019autodispnet,chen2019fasterseg,esmaeilzehi2021fpnet}.

Third, most NAS studies tied architecture search to a single task, either semantic segmentation~\cite{chen2019fasterseg} or depth estimation~\cite{saikia2019autodispnet,huynh2021lightweight} or image super-resolution~\cite{esmaeilzehi2021fpnet}. By utilizing ATS together with a generic pre-defined backbone, LDP generalizes architecture search for various dense prediction problems.

Figure~\ref{fig:rel_params_chartv1} presents a comparison between our LDP models and
other state-of-the-art lightweight approaches for monocular depth estimation, semantic segmentation and image super-resolution tasks. Compared to PyD-Net~\cite{poggi2018towards}, our method improves the REL, RMSE, and thresholded accuracy by $13.6\%$, $8.3\%$, and $3\%$ with similar execution time on the Google Pixel 3a phone (see Table~\ref{tab:runtime_comparison}). Compared to FastDepth~\cite{wofk2019fastdepth} and EDA~\cite{tu2020efficient}, our model achieves higher accuracy with fewer parameters.

To summarize, our work makes the following contributions:
\begin{itemize}[noitemsep,topsep=3pt,parsep=3pt,partopsep=3pt]
\item We propose a multi-objective exploration framework, LDP, searching for accurate and lightweight dense prediction architectures. It extends our previous work~\cite{huynh2021lightweight} on monocular depth estimation to multiple prediction problems.
\item We leverage the novel Assisted Tabu Search to enable fast neural architecture search.
\item We create a well-defined search space that allows computational flexibility and layer diversity.
\item We achieve state-of-the-art results compared to lightweight baselines on diverse problems and datasets, with 
significantly less parameters.
\end{itemize}
\noindent The implementation of LDP will be made publicly available upon publication of the paper.

\begin{figure*}[!t]
\begin{center}
  \includegraphics[width=0.9\linewidth]{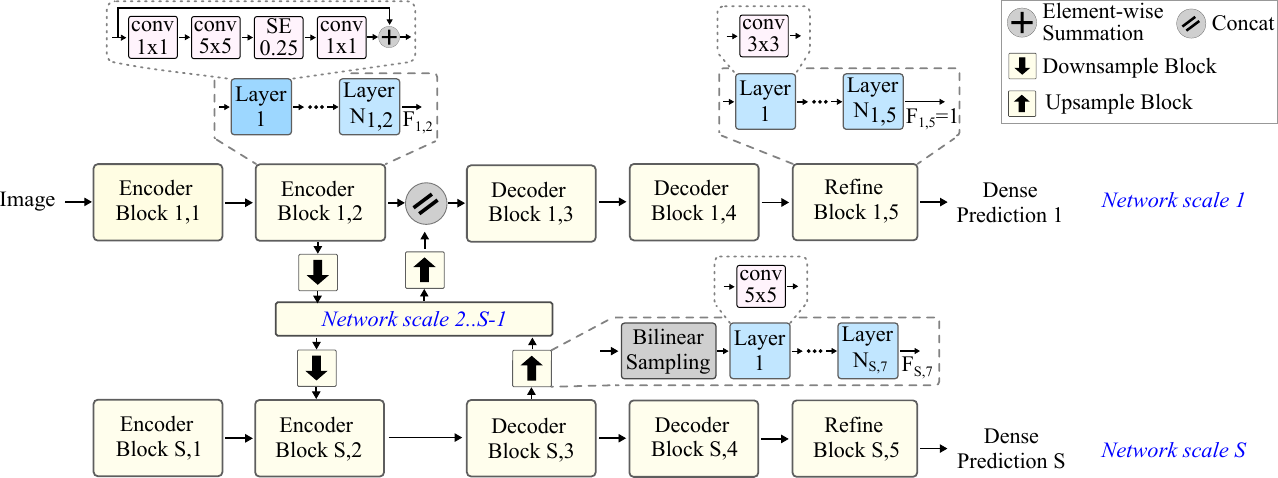}
\end{center}
\vspace{-0.35cm}
  \caption{The search space of our LDP framework.  Models are constructed from a pre-defined backbone network containing encoder, decoder, refine, downsample and upsample \textit{blocks} (green).  A block is formed by several identical layers (orange) that are generated from a pool of operations and connections. Layers within a block are the same while layers of different blocks can be different.} \vspace{-0.45cm}
\label{fig:search_space_ver2}
\end{figure*}

\section{Related work}
\noindent \textbf{Dense prediction problem} We focus on three dense prediction subtasks: such as monocular depth estimation, semantic segmentation and image super-resolution.

Learning-based monocular depth estimation was first introduced by Saxena et al.~\cite{saxena2006learning}. Later studies improved accuracy by using large network architectures~\cite{chen2019structure,eigen2015predicting,eigen2014depth,Hu2018RevisitingSI,laina2016deeper} or integrating semantic information~\cite{jiao2018look} and surface normals~\cite{qi2018geonet}. Fu et al.~\cite{fu2018deep} formulated depth estimation as an ordinal regression problem, while~\cite{chen2016single,lee2019monocular} estimated relative instead of metric depth.  Facil et al.~\cite{facil2019cam} proposed to learn camera calibration from the images for depth estimation. Recent approaches further improve the performance by exploiting monocular priors such as planarity constraints~\cite{liu2019planercnn,liu2018planenet,Yin2019enforcing,huynh2020guiding,lee2019big} or occlusion~\cite{ramamonjisoa2019sharpnet}. Gonzalez and Kim~\cite{gonzalezbello2020forget} estimated depth by synthesizing stereo pairs from a single image, while~\cite{yang2021transformers} and~\cite{Ranftl2021} applied vision-transformer for depth prediction. For resource-limited hardware, it is more desirable to not only have a fast but also accurate model. A simple alternative is employing lightweight architectures such as MobileNet~\cite{howard2019searching,howard2017mobilenets,sandler2018mobilenetv2,wofk2019fastdepth}, GhostNet~\cite{han2020ghostnet}, and FBNet~\cite{tu2020efficient}. A popular approach is utilizing network compression techniques, including quantization~\cite{han2015deep}, network pruning~\cite{yang2018netadapt}, knowledge distillation~\cite{yucel2021real}. Another approach employs well-known pyramid networks or dynamic optimization schemes~\cite{aleotti2021real}.

Deep neural networks have thrived on semantic segmentation, with early works~\cite{girshick2014rich,hariharan2014simultaneous} proposed to classify region proposals to perform this task. Later on, fully convolutional neural network methods~\cite{long2015fully,dai2015boxsup} are widely adopted to process arbitrary-sized input images and train the network end-to-end. Atrous convolution-based approach~\cite{liang2015semantic} generated the middle-score map that was later refined using the dense conditional random field (CRF)~\cite{krahenbuhl2011efficient} to mitigate the low-resolution prediction problem. Chen et al.~\cite{chen2017deeplab} then implemented atrous spatial pyramid pooling for segmenting objects at different scales, while~\cite{chen2017rethinking} and~\cite{chen2018encoder} employed atrous separable convolution and an efficient decoder module to capture sharp object boundaries. Zheng et al.~\cite{zheng2015conditional} enabled end-to-end training of dense CRF by implementing recurrent layers. Other approaches~\cite{badrinarayanan2017segnet,noh2015learning} utilized transposed convolution to generate the high-resolution prediction. Long et al.~\cite{long2015fully} produced multi-resolution prediction scores and took the average to generate the final output. Hariharan et al.~\cite{hariharan2015hypercolumns} fused mid-level features and trained dense classification layers at multiple stages. Badrinarayanan et al.~\cite{badrinarayanan2017segnet} and Ronneberger et al.~\cite{ronneberger2015u} implemented transposed convolution with skip-connections to exploit mid-level features. Wang et al.~\cite{wang2020deep} utilized multi-scale parallel inter-connected layers to further exploit learned features from pre-trained ImageNet models. Lightweight approaches~\cite{orsic2019defense,li2019dfanet,yu2018bisenet} employed pre-trained backbones as a decoder and a simple decoder to perform fast segmentation. Zhao et al.~\cite{zhao2018icnet} modified the cascade architecture of~\cite{zhao2017pyramid} to shrink the model size and speed-up inference.

Image super-resolution task has also been immensely improved using deep neural networks. Dong et al.~\cite{dong2015image} first proposed a shallow but bulky network that~\cite{dong2016accelerating} later shrunk down utilizing transposed convolution. Kim et al.~\cite{kim2016accurate} implemented a deeper architecture with skip connection while Zhang et al.~\cite{zhang2018image} proposed channel attention to improve the performance. Kim et al.~\cite{kim2016deeply} introduced recursive layers that Tai et al.~\cite{tai2017image} later extended by adding the local residual connection. Likewise, many studies attempted to enhance model efficiency. Ahn et al.~\cite{ahn2018fast} suggested using cascade architecture and group convolution to reduce the number of parameters. Hui et al.~\cite{hui2018fast,hui2019lightweight} proposed the information distillation module for creating an efficient model. Lai et al.~\cite{lai2017deep} applied the Laplacian pyramid network to gradually increase the spatial resolution while downsizing the model.

However, state-of-the-art methods mainly focus on increasing accuracy at the cost of model complexity that is infeasible in resource-limited settings, while manually designed lightweight architecture is a tedious task, requires much trial-and-error, and usually leads to architectures with low accuracy.

\noindent \textbf{Neural Architecture Search} There has been increasing interest in automating network design using neural architecture search. Most of these methods focus on searching high-performance architecture using reinforcement learning~\cite{baker2016designing,liu2018progressive,pham2018efficient,zoph2016neural,zoph2018learning}, evolutionary search~\cite{real2019regularized}, differentiable search~\cite{liu2018darts}, or other learning algorithms~\cite{luo2018neural}. However, these methods are usually very slow and require huge resources for training. Other studies~\cite{dong2018dpp,elsken2018multi,hsu2018monas} also attempt to optimize multiple objectives like model size and accuracy. Nevertheless, their search
process optimizes only on small tasks like CIFAR. In contrast,
our proposed method targets real-world data such as NYU, KITTI, and Cityscapes on multiple dense prediction tasks.

\section{Lightweight Dense Precition (LDP)}

\noindent We propose the LDP framework to search for accurate and lightweight monocular depth estimation architectures utilizing a pre-defined backbone that has been successful for dense prediction in the past.
The proposed framework takes in a dataset as input to search for the best possible model.  This model can be deployed for depth estimation on hardware-limited devices. The first subsection defines the search space while the remaining two describe our multi-objective exploration and search algorithm.

\subsection{Search Space}

\noindent Previous neural architecture search (NAS) studies demonstrated the significance of designing a well-defined search space. A common choice of NAS is searching for a small set of complicated cells from a smaller dataset~\cite{zoph2018learning,liu2018progressive,real2019regularized}. These cells are later replicated to construct the entire architecture that hindered layer diversity and suffered from domain differences~\cite{tan2019mnasnet}. On the other hand, unlike classification tasks, dense prediction problems involve mapping a feature representation in the encoder to predictions at larger spatial resolution in the decoder.

To this end, we build our search space upon a pre-defined backbone that is shown as the set of green blocks in Figure~\ref{fig:search_space_ver2}. The backbone is divided into multi-scale pyramid networks operating at different spatial resolutions. Each network scale consists of two encoder blocks, two decoder blocks, a refinement block, a downsampling and upsampling block (except for scale 1). Each block is constructed from a set of identical layers (marked as orange in Figure~\ref{fig:search_space_ver2}). Inspired by~\cite{tan2019mnasnet}, we search for the layer from a pool of operations and connections, including:

\begin{figure*}[!t]
\begin{center}
  \includegraphics[width=0.99\linewidth]{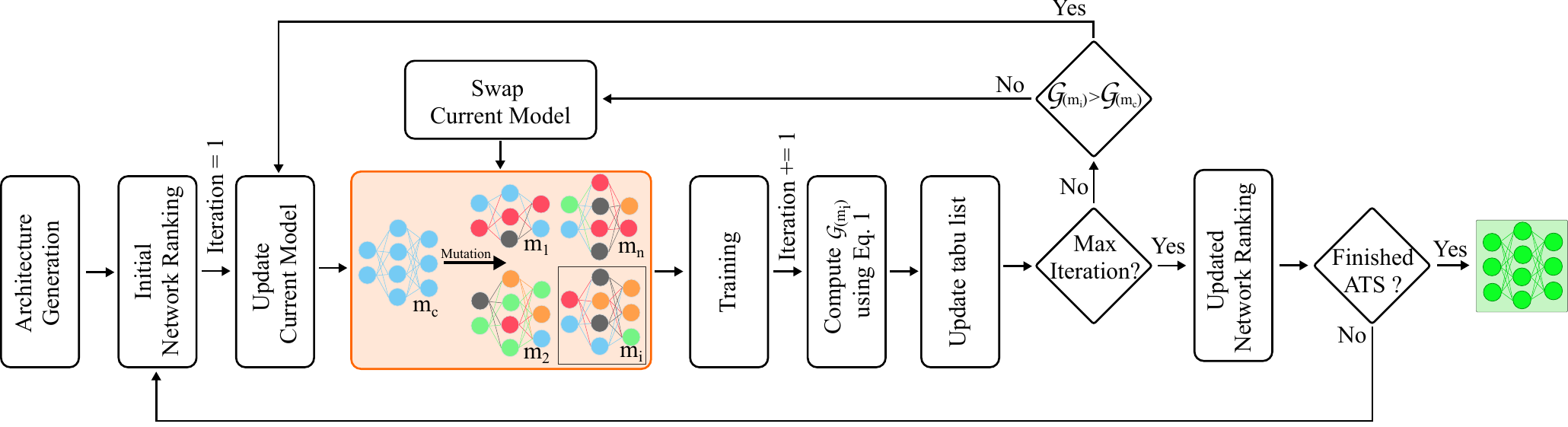}
\end{center}
\vspace{-0.35cm}
  \caption{The flowchart of our architecture search that utilizes the Assisted Tabu Search (ATS) with mutation to search for accurate and lightweight monocular depth estimation networks.} \vspace{-0.55cm}
\label{fig:search_algorithm}
\end{figure*}

\begin{itemize}[noitemsep,topsep=0pt,parsep=1pt,partopsep=1pt]
    \item The number of resolution scales $S$.
    \item The number of layers for each block $N_{i,j}$.
    \item Convolutional operations (ConvOps): vanilla 2D convolution, depthwise convolution, inverted bottleneck convolution and micro-blocks~\cite{li2021micronet}.
    \item Convolutional kernel size (KSize): $3 \times 3$, $5 \times 5$.
    \item Squeeze and excitation ratio (SER): $0$, $0.25$.
    \item Skip connections (SOps): residual or no connection.
    \item The number of output channels: $F_{i,j}$.
\end{itemize}

\noindent Here $i$ indicates the resolution scale and $j$ is the block index at the same resolution. Internal operations such as ConvOps, KSize, SER, SOps, $F_{i,j}$ are utilized to construct the layer while $N_{i,j}$ determines the number of layers that will be replicated for block$_{i,j}$. In other words, as shown in Figure~\ref{fig:search_space_ver2}, layers within a block (e.g. layers 1 to N$_{1,2}$ of Encoder Block 1,2 are the same) are similar while layers of different blocks (e.g. Layer 1 in Refine Block 1,5 versus Layer 1 in Upsample Block S,7) can be different.

We also perform layer mutation to further diversifying the network structure during the architecture search process. The mutation operations include:
\begin{itemize}[noitemsep,topsep=0pt,parsep=1pt,partopsep=1pt]
    \item Swapping operations of two random layers with compatibility check.
    \item Modifying a layer with a new valid layer from the predefined operations.
\end{itemize}

Moreover, we also set computational constraints to balance the kernel size with the number of output channels. Therefore, increasing the kernel size of one layer usually results in decreasing output channels of another layer. 

Assuming we have a network of $S$ scales, and each block has a sub-search space of size $M$ then our total search space will be $M^{5 + [(S - 1) * 7]}$. Supposedly, a standard case with $M=192$, $S=5$ will result in a search space of size $\sim 2 \times 10^{75}$.

\subsection{Multi-Objective Exploration}

\noindent We introduce a multi-objective search paradigm seeking for both accurate and compact architectures. For this purpose, we monitor the \textit{validation grade} $\mathcal{G}$ that formulates both accuracy and the number of parameter of the trained model. It is defined by

\begin{equation} 
\label{eq:validation_grade}
\mathcal{G}(m) = \alpha \times A(m) + (1 - \alpha) \times \bigg[\frac{P}{P(m)}\bigg]^{r}
\end{equation}

\noindent where $A(m)$ and $P(m)$ are validation accuracy and the number of parameters of model $m$. $P$ is the target compactness, $\alpha$ is the balance coefficient, and $r$ is an exponent with $r=0$ when $P(m) \leq P$ and otherwise $r=1$.  The goal is to search for an architecture $m^{*}$ where $G(m^{*})$ is maximum.

However, computing $G$ requires training for every architecture candidate, resulting in considerable search time. To mitigate this problem, Mellor et al.~\cite{mellor2021neural} suggested to score an architecture at initialisation to predict its performance before training. For a network $f$, the \textit{score(f)} is defined as:

\begin{equation} 
\label{eq:jacob_cov_score}
score(f) = log|K_{H}|
\end{equation}

\noindent where $K_{H}$ is the kernel matrix. 
Let us assume that the mapping of model $f$ from a batch of data $X = \{x_i\}^{N}_{i=1}$ is $f(x_i)$. 
By assigning binary indicators to every activation units in $f$, a linear region $x_i$ of data point $i$ is represented by the binary code $c_i$. The kernel matrix $K_{H}$ is defined as:

\begin{equation} 
\label{eq:kernel_matrix}
K_{H} = \begin{pmatrix}
N_{A} - d_{H}(c_1, c_1) & \dots & N_{A} - d_{H}(c_1, c_N) \\
\vdots & \ddots & \vdots \\
N_{A} - d_{H}(c_N, c_1) & \dots & N_{A} - d_{H}(c_N, c_N)
\end{pmatrix}
\end{equation}

\noindent where $N_A$ is the number of activation units, and $d_{H}(c_i, c_j)$ is the Hamming distance between two binary codes. Inspired by this principle, we generate and train a set of different architectures for various dense prediction tasks. We evaluate the performance of these models and visualize the results against the \textit{score} that in our case is the mapping of depth values within image batches. 
Leveraging this observation, we 1) utilize the \textit{score} in our initial network ranking, and 2) define the mutation exploration reward $\mathcal{R}$ as:

\begin{equation} 
\label{eq:mutation_exploration_reward}
\mathcal{R}(m_i,m_j) = \alpha \times \frac{score(m_j)}{score(m_i)} + (1 - \alpha) \times \bigg[\frac{P}{P(m_j)}\bigg]^{r}
\end{equation}

\noindent where $m_j$ is a child network that is mutated from $m_i$ architecture.

\subsection{Search Algorithm}

\noindent The flowchart of our architecture search is presented in Figure~\ref{fig:search_algorithm}. We first randomly generate $60K$ unique parent models and create the initial network ranking based on the \textit{score} in Eq.~\ref{eq:jacob_cov_score}. We then select \textit{six} architectures in which \textit{three} are the highest-ranked while the other \textit{three} have the highest score of the networks with the size closest to the target compactness. 

\begin{table*}[t!]
\caption{\label{tab:eval_nyuv2}Evaluation on the NYU-Depth-v2 dataset. Metrics with $\downarrow$ mean lower is better and $\uparrow$ mean higher is better. Type column shows the exploration method used to obtain the model. RL, ATS, and manual, refer to reinforcement learning, assisted tabu search, and manual design, respectively.}
\centering
\small
\begin{tabular}{@{}llrcccccccc@{}}
\hline
\multicolumn{2}{c}{\textbf{Architecture}} & \textbf{\#params} & \textbf{Type} & \textbf{Search Time} & \textbf{REL$\downarrow$} & \textbf{RMSE$\downarrow$} & \(\boldsymbol{\delta_{1}}\)$\uparrow$ & \(\boldsymbol{\delta_{2}}\)$\uparrow$ & \(\boldsymbol{\delta_{3}}\)$\uparrow$ \\ \hline

AutoDepth-BOHB-S & Saikia et al.'19~\cite{saikia2019autodispnet} & 63.0M & RL & 42 GPU days & 0.170 & 0.599 & - & - & - \\ \hline

EDA & Tu et al.'21~\cite{tu2020efficient} & 5.0M & Manual & - & 0.161 & 0.557 & 0.782 & 0.945 & 0.984 \\ \hline

Ef+FBNet & Tu \& Wu et al.~\cite{tu2020efficient,wu2019fbnet}  & 4.7M & Manual & - & 0.149 & 0.531 & 0.803 & 0.952 & 0.987 \\ \hline

FastDepth & Wofk et al.'19~\cite{wofk2019fastdepth} & 3.9M & Manual & - & 0.155 & 0.599 & 0.778 & 0.944 & 0.981 \\ \hline

SparseSupNet & Yucel et al.'21~\cite{yucel2021real} & 2.6M & Manual & - & 0.153 & 0.561 & 0.792 & 0.949 & 0.985 \\ \hline

LiDNAS-N & Huynh et al.'21~\cite{huynh2021lightweight}  & 2.1M & ATS & 4.3 GPU days & \textbf{0.132} & 0.487 & 0.845 & 0.965 & \textbf{0.993} \\ \hline

LDP-Depth-N & Ours  & \textbf{2.0M} & ATS & 4.3 GPU days & \textbf{0.132} & \textbf{0.483} & \textbf{0.848} & \textbf{0.967} & \textbf{0.993} \\ \hline

\end{tabular} \vspace{-0.45cm}
\end{table*}

Starting from these initial networks, we strive for the best possible model utilizing Assisted Tabu Search (ATS), inspired by Tabu search (TS)~\cite{glover1986future} that is a high level procedure for solving multicriteria optimization problems. It is an iterative algorithm that starts from some initial feasible solutions and aims to determine better solutions while being designed to avoid traps at local minima.

We propose ATS by applying Eq.~\ref{eq:validation_grade} and~\ref{eq:mutation_exploration_reward} to TS to speed up the searching process. Specifically, we mutate numerous children models ($m_1$, $m_2$, .., $m_n$) from the current architecture ($m_c$). The mutation exploration reward $\mathcal{R}(m_c, m_i)$ is calculated using Eq.~\ref{eq:mutation_exploration_reward}.
ATS then chooses to train the mutation with the highest rewards (e.g. architecture $m_i$ as demonstrated in Figure~\ref{fig:search_algorithm}). The validation grade of this model $\mathcal{G}(m_i)$ is calculated after the training. The performance of the chosen model is assessed by comparing $\mathcal{G}(m_i)$ with $\mathcal{G}(m_c)$. If $\mathcal{G}(m_i)$ is larger than $\mathcal{G}(m_c)$, then $m_i$ is a good mutation, and we opt to build the next generation upon its structure. Otherwise, we swap to use the best option in the tabu list for the next mutation. The process stops when reaching a maximum number of iterations or achieving a terminal condition. The network ranking will be updated, and the search will continue for the remaining parent architectures.

\subsection{Implementation Details} \label{implementation_detail}

\noindent For searching, we directly perform our architecture exploration on the training samples of the target dataset. We set the target compactness parameter $P$ using the previously published compact models as a guideline. We set the maximum number of exploration iterations to 100 and stop the exploration procedure if a better solution cannot be found after 10 iterations. The total search time required to find optimal architecture is $\sim 4.3$ GPU days.

\begin{table}[b!]
\caption{\label{tab:eval_kitti}Evaluation on the KITTI dataset. Metrics with $\downarrow$ mean lower is better and $\uparrow$ mean higher is better.}
\centering
\small
\adjustbox{width=\columnwidth}{\begin{tabular}{@{}lrccccc@{}}
\hline
\textbf{Method} &\textbf{\#params} & \textbf{REL$\downarrow$} & \textbf{RMSE$\downarrow$} & \(\boldsymbol{\delta_{1}}\)$\uparrow$ & \(\boldsymbol{\delta_{2}}\)$\uparrow$ & \(\boldsymbol{\delta_{3}}\)$\uparrow$ \\ \hline

FastDepth~\cite{wofk2019fastdepth} & 3.93M & 0.156 & 5.628 & 0.801 & 0.930 & 0.971 \\ \hline

PyD-Net~\cite{poggi2018towards} & 1.97M & 0.154 & 5.556 & 0.812 & 0.932 & 0.970 \\ \hline

EQPyD-Net~\cite{cipolletta2021energy} & 1.97M & 0.135 & 5.505 & 0.821 & 0.933 & 0.970 \\ \hline

DSNet~\cite{aleotti2021real} & 1.91M & 0.159 & 5.593 & 0.800 & 0.932 & 0.971 \\ \hline

LiDNAS-K & 1.78M & \textbf{0.133} & 5.157 & 0.842 & \textbf{0.948} & 0.980 \\ \hline

LDP-Depth-K & \textbf{1.74M} & \textbf{0.133} & \textbf{5.155} & \textbf{0.844} & \textbf{0.948} & \textbf{0.981} \\ \hline
\end{tabular}} \vspace{-0.25cm}
\end{table}

\begin{table}[b!]
\vspace{-0.15cm}
\caption{\label{tab:eval_cityscapes}Segmentation results on Cityscapes dataset. (MIoU)}
\centering
\small
\begin{tabular}{@{}lrccc@{}}
\hline
{\textbf{Model}} & \textbf{\#params} & \textbf{resolution} & \textbf{val}$\uparrow$ & \textbf{test}$\uparrow$ \\ \hline

BiSeNetV1 B~\cite{yu2018bisenet} & 49.0M & $768 \times 1536$ & 74.8 & 74.7 \\ \hline

SwiftNet\cite{orsic2019defense} & 11.8M & $512 \times 1024$ & 70.2 & - \\ \hline

DFANet~\cite{li2019dfanet} & 7.8M & $512 \times 1024$ & 70.8 & 70.3 \\ \hline

BiSeNetV1 A~\cite{yu2018bisenet} & 5.8M & $768 \times 1536$ & 69.0 & 68.4 \\ \hline

MobileNeXt~\cite{zhou2020rethinking} & 4.5M & $1024 \times 2048$ & 75.5 & 75.2 \\ \hline

BiSeNetV2~\cite{yu2021bisenet} & 4.3M & $1024 \times 2048$ & 75.8 & 75.3 \\ \hline

HRNet-W16~\cite{wang2020deep} & 2.0M & $512 \times 1024$ & 68.6 & 68.1 \\ \hline 

Lite-HRNet~\cite{yu2021lite} & 1.8M & $512 \times 1024$ & \textbf{76.0} & 75.3 \\ \hline  \hline

FasterSeg~\cite{chen2019fasterseg} & 4.4M & $1024 \times 2048$ & 73.1 & 71.5 \\ \hline

MobileNetV3~\cite{howard2019searching} & \textbf{1.5M} & $1024 \times 2048$ & 72.4 & 72.6 \\ \hline

LDP-Seg-Ci & 1.7M & $512 \times 1024$ & 75.8 & \textbf{75.5} \\ \hline \hline

\end{tabular} 
\vspace{-0.25cm}
\end{table}

\begin{table}[b!]
\caption{\label{tab:eval_coco}Segmentation results on COCO-Stuff dataset.}
\centering
\small
\begin{tabular}{@{}lrccc@{}}
\hline
{\textbf{Model}} & \textbf{\#params} & \textbf{PixAcc}(\%)$\uparrow$ & \textbf{MIoU}(\%)$\uparrow$  \\ \hline

BiSeNetV1 B~\cite{yu2018bisenet} & 49.0M & 63.2 & 28.1 \\ \hline

ICNet~\cite{zhao2018icnet} & 12.7M & - & 29.1 \\ \hline

BiSeNetV1 A~\cite{yu2018bisenet} & 5.8M & 59.0 & 22.8 \\ \hline

BiSeNetV2~\cite{yu2021bisenet} & \textbf{4.3M} & 63.5 & 28.7 \\ \hline

LDP-Seg-CO & \textbf{4.3M} & \textbf{64.2} & \textbf{29.3} \\ \hline \hline
\end{tabular} 
\end{table}

For training, we use the Adam optimizer~\cite{kingma2014adam} with $(\beta_1, \beta_2, \epsilon) = (0.9, 0.999, 10^{-8})$. The initial learning rate is $7*10^{-4}$, but from epoch 10 the learning is reduced by $5\%$ per $5$ epochs. We use a batch size of 256 and augment the input RGB and ground truth depth images using random rotations ([-5.0, +5.0] degrees), horizontal flips, rectangular window droppings, and colorization (RGB only).

\section{Experiments}
We deploy the LDP framework on dense prediction tasks: monocular depth estimation, semantic segmentation, and image super-resolution. Experiments show that LDP improved performance while using only a fraction of the number of parameters needed by the competing approaches.

\subsection{Monocular Depth Estimation}
We first demonstrate our method for monocular depth estimation that can be formulated as a dense regression problem. The main goal is to infer continuous pixel-wise depth values from a single input image. For this task, we apply LDP on the NYU-Depth-v2 \cite{silberman2012indoor} and KITTI~\cite{geiger2013vision} datasets. NYU-Depth-v2 contains $\sim120K$ RGB-D images obtained from 464 indoor scenes. From the entire dataset, we use 50K images for training and the official test set of 654 images for evaluation. KITTI is an outdoor driving dataset, where we use the standard Eigen split~\cite{eigen2015predicting,eigen2014depth} for training (39K images) and testing (697 images). We report the mean absolute relative error (REL), root mean square error (RMSE), and thresholded accuracy ($\delta_i$) as our performance metrics.

\begin{table*}[t!]
\caption{\label{tab:sup_res_x2}Image Super-Resolution with scaling factor $\times 2$.} 
\centering
\begin{tabular}{@{}lccccccccc@{}}
\hline
\multicolumn{1}{c}{\multirow{2}{*}{ \textbf{Method} }} &
\multicolumn{1}{c}{\multirow{2}{*}{ \textbf{\#params} }} &
  \multicolumn{2}{c}{ \textbf{Set5} } &
  \multicolumn{2}{c}{ \textbf{Set14} } &
  \multicolumn{2}{c}{ \textbf{BSD100} } &
  \multicolumn{2}{c}{ \textbf{Urban100} } \\ \cmidrule(l){3-10} 
\multicolumn{2}{c}{} & PSNR$\uparrow$ & SSIM$\uparrow$ & PSNR$\uparrow$ & SSIM$\uparrow$ & PSNR$\uparrow$ & SSIM$\uparrow$ & PSNR$\uparrow$ & SSIM$\uparrow$  \\ \midrule
Bicubic & - & 33.66 & 0.9299 & 30.24 & 0.8688 & 29.56 & 0.8431 & 26.91 & 0.8425 \\
CARN~\cite{ahn2018fast} & 1.59M & 37.76 & 0.9590 & 33.52 & 0.9166 & 32.09 & 0.8978 & 31.92 & 0.9256 \\
LFFN~\cite{yang2019lightweight} & 1.52M & 37.95 & 0.9597 & 32.45 & 0.9142 & 32.20 & 0.8994 & 32.39 & 0.9299 \\
OISR-LF-s~\cite{he2019ode} & 1.37M & 38.02 & 0.9605 & 33.62 & 0.9178 & 32.20 & 0.9000 & 32.21 & 0.9290 \\
CBPN~\cite{zhu2019efficient} & 1.04M & 37.90 & 0.9590 & 33.60 & 0.9171 & 32.17 & 0.8989 & 32.14 & 0.9279 \\
OverNet~\cite{behjati2021overnet} & 0.90M & 38.11 & 0.9610 & 33.71 & 0.9179 & 32.24 & 0.9007 & 32.44 & 0.9311 \\
LapSRN~\cite{lai2017deep} & 0.81M & 37.52 & 0.9590 & 33.08 & 0.9130 & 31.80 & 0.8950 & 30.41 & 0.9100 \\
IMDN~\cite{hui2019lightweight} & 0.69M & 38.00 & 0.9605 & 33.63 & 0.9177 & 32.19 & 0.8996 & 32.17 & 0.9283 \\
MemNet~\cite{tai2017memnet} & 0.67M & 37.78 & 0.9597 & 33.28 & 0.9142 & 32.08 & 0.8978 & 31.31 & 0.9195 \\
VDSR~\cite{kim2016accurate} & 0.66M & 37.53 & 0.9587 & 33.05 & 0.9127 & 31.90 & 0.8960 & 30.77 & 0.9141 \\
IDN~\cite{hui2018fast} & 0.57M & 37.85 & 0.9598 & 33.58 & 0.9178 & 32.11 & 0.8989 & 31.95 & 0.9266 \\
CARN-M~\cite{ahn2018fast} & 0.41M & 37.53 & 0.9583 & 33.26 & 0.9141 & 31.92 & 0.8960 & 31.23 & 0.9194 \\
DRRN~\cite{tai2017image} & 0.29M & 37.74 & 0.9591 & 33.23 & 0.9136 & 32.05 & 0.8973 & 31.23 & 0.9188 \\
SRFBN~\cite{li2019feedback} & 0.28M & 37.78 & 0.9597 & 33.35 & 0.9156 & 32.00 & 0.8970 & 31.41 & 0.9207 \\
FSRCNN~\cite{dong2016accelerating} & \textbf{0.12M} & 37.00 & 0.9558 & 32.63 & 0.9088 & 31.53 & 0.8920 & 29.88 & 0.9020 \\ \hline

DeCoNASNet~\cite{ahn2021neural} & 1.71M & 37.96 & 0.9594 & 33.63 & 0.9175 & 32.15 & 0.8986 & 32.03 & 0.9265 \\
FPNet~\cite{esmaeilzehi2021fpnet} & 1.61M & \textbf{38.13} & \textbf{0.9616} & \textbf{33.83} & \textbf{0.9198} & \textbf{32.29} & 0.9018 & 32.04 & 0.9278 \\
FALSR-A~\cite{chu2021fast} & 1.03M & 37.82 & 0.9595 & 33.55 & 0.9168 & 32.12 & 0.8987 & 31.93 & 0.9256 \\
LDP-Sup-x2 & 1.02M & 38.11 & 0.9612 & \textbf{33.83} & 0.9196 & \textbf{32.29} & \textbf{0.9019} & \textbf{32.49} & \textbf{0.9314} \\
\hline \hline
\end{tabular}
\end{table*}

\begin{figure*}[ht]
\begin{center}
  \includegraphics[width=0.99\linewidth]{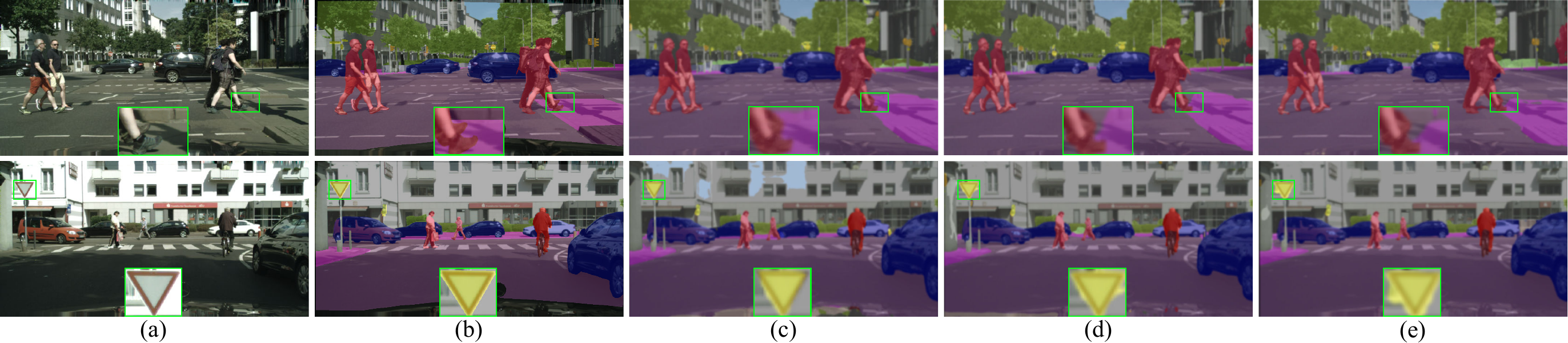}
\end{center}
\vspace{-0.35cm}
  \caption{Comparison on the Cityscapes validation set. 
  (a) input image,
  (b) ground truth,
  (c) LDP-Seg-Ci,
  (d) Lite-HRNet~\cite{yu2021lite},
  and (e) MobileNetV3~\cite{howard2019searching}.}
\label{fig:qualitative_cityscapes} \vspace{-0.35cm}
\end{figure*}

In the case of NYU-Depth-v2, we set the target compactness $P=1.8M$ with the balance coefficient $\alpha=0.6$ to search for the optimized model on NYU-Depth-v2. We then select the best performance model (LDP-Depth-N) and compare its results with lightweight state-of-the-art methods~\cite{tu2020efficient,wofk2019fastdepth,wu2019fbnet,yucel2021real} along with their numbers of parameters. As shown in Table~\ref{tab:eval_nyuv2}, LDP-Depth-N outperforms the baseline while containing the least amount of parameters. Comparing with the best-performing approach, the proposed model improves the REL, RMSE, and $\theta_{1}$ by $11.4\%$, $8.2\%$, and $6.8\%$ while compressing the model size by $55\%$. Our method produces high-quality depth maps with sharper details as presented in Figure~\ref{fig:qualitative_nyu}. However, we observe that all methods still struggle in challenging cases, such as the scene containing Lambertian surfaces as illustrated by the example in the third column of Figure~\ref{fig:qualitative_nyu}. Moreover, the proposed method improves REL and RMSE by $22.3\%$ and $18.7\%$ while using only $3\%$ of the model parameters comparing to the state-of-the-art NAS-based disparity and depth estimation approaches~\cite{saikia2019autodispnet}. In addition, our method requires $90\%$ less search time than~\cite{saikia2019autodispnet} and outperforms~\cite{huynh2021lightweight} in almost all metrics.

\begin{table*}[t!]
\caption{\label{tab:sup_res_x4}Image Super-Resolution with scaling factor $\times 4$.} 
\centering
\begin{tabular}{@{}lccccccccc@{}}
\hline
\multicolumn{1}{c}{\multirow{2}{*}{ \textbf{Method} }} &
\multicolumn{1}{c}{\multirow{2}{*}{ \textbf{\#params} }} &
  \multicolumn{2}{c}{ \textbf{Set5} } &
  \multicolumn{2}{c}{ \textbf{Set14} } &
  \multicolumn{2}{c}{ \textbf{BSD100} } &
  \multicolumn{2}{c}{ \textbf{Urban100} } \\ \cmidrule(l){3-10} 
\multicolumn{2}{c}{} & PSNR$\uparrow$ & SSIM$\uparrow$ & PSNR$\uparrow$ & SSIM$\uparrow$ & PSNR$\uparrow$ & SSIM$\uparrow$ & PSNR$\uparrow$ & SSIM$\uparrow$  \\ \midrule
Bicubic & - & 28.42 & 0.8104 & 26.01 & 0.7027 & 25.96 & 0.6675 & 23.17 & 0.6585 \\
s-LWSR64~\cite{li2020s} & 2.27M & 32.28 & 0.8960 & 28.34 & 0.7800 & 27.61 & 0.7380 & 26.19 & 0.7910 \\
CARN~\cite{ahn2018fast} & 1.59M & 32.13 & 0.8937 & 28.60 & 0.7806 & 27.58 & 0.7349 & 26.07 & 0.7837 \\
LFFN~\cite{yang2019lightweight} & 1.53M & 32.15 & 0.8945 & 28.32 & 0.7810 & 27.52 & 0.7377 & 26.24 & 0.7902 \\
OISR-LF-s~\cite{he2019ode} & 1.52M & 32.14 & 0.8947 & 28.63 & 0.7819 & 27.60 & 0.7369 & 26.17 & 0.7888 \\
CBPN~\cite{zhu2019efficient} & 1.19M & 32.21 & 0.8944 & 28.63 & 0.7813 & 27.58 & 0.7356 & 26.14 & 0.7869 \\
LapSRN~\cite{lai2017deep} & 0.81M & 31.54 & 0.8850 & 28.19 & 0.7720 & 27.32 & 0.7280 & 25.21 & 0.7560 \\
IMDN~\cite{hui2019lightweight} & 0.72M & 32.21 & 0.8948 & 28.58 & 0.7811 & 27.56 & 0.7353 & 26.04 & 0.7838 \\
MemNet~\cite{tai2017memnet} & 0.67M & 31.74 & 0.8893 & 28.26 & 0.7723 & 27.40 & 0.7281 & 25.50 & 0.7630 \\
VDSR~\cite{kim2016accurate} & 0.66M & 31.33 & 0.8838 & 28.02 & 0.7678 & 27.29 & 0.7252 & 25.18 & 0.7525 \\
IDN~\cite{hui2018fast} & 0.60M & 31.99 & 0.8928 & 28.52 & 0.7794 & 27.52 & 0.7339 & 25.92 & 0.7801 \\
s-LWSR32~\cite{li2020s} & 0.57M & 32.04 & 0.8930 & 28.15 & 0.7760 & 27.52 & 0.7340 & 25.87 & 0.7790 \\
SRFBN~\cite{li2019feedback} & 0.48M & 31.98 & 0.8923 & 28.45 & 0.7779 & 27.44 & 0.7313 & 25.71 & 0.7719 \\
CARN-M~\cite{ahn2018fast} & 0.41M & 31.92 & 0.8903 & 28.42 & 0.7762 & 27.44 & 0.7304 & 25.62 & 0.7694 \\
DRRN~\cite{tai2017image} & 0.29M & 31.68 & 0.8888 & 28.21 & 0.7720 & 27.38 & 0.7284 & 25.44 & 0.7638 \\
FSRCNN~\cite{dong2016accelerating} & \textbf{0.12M} & 30.71 & 0.8657 & 27.59 & 0.7535 & 26.98 & 0.7150 & 24.62 & 0.7280 \\ \hline

FPNet~\cite{esmaeilzehi2021fpnet} & 1.61M & \textbf{32.32} & 0.8962 & \textbf{28.78} & \textbf{0.7856} & 27.66 & \textbf{0.7394} & 26.09 & 0.7850 \\
LDP-Sup-x4 & 1.09M & 32.30 & \textbf{0.8963} & 28.54 & 0.7836 & \textbf{27.67} & \textbf{0.7394} & \textbf{26.25} & \textbf{0.7927} \\
\hline \hline
\end{tabular} 
\end{table*} \vspace{-0.15cm} 

\begin{figure*}[ht]
\begin{center}
  \includegraphics[width=0.99\linewidth]{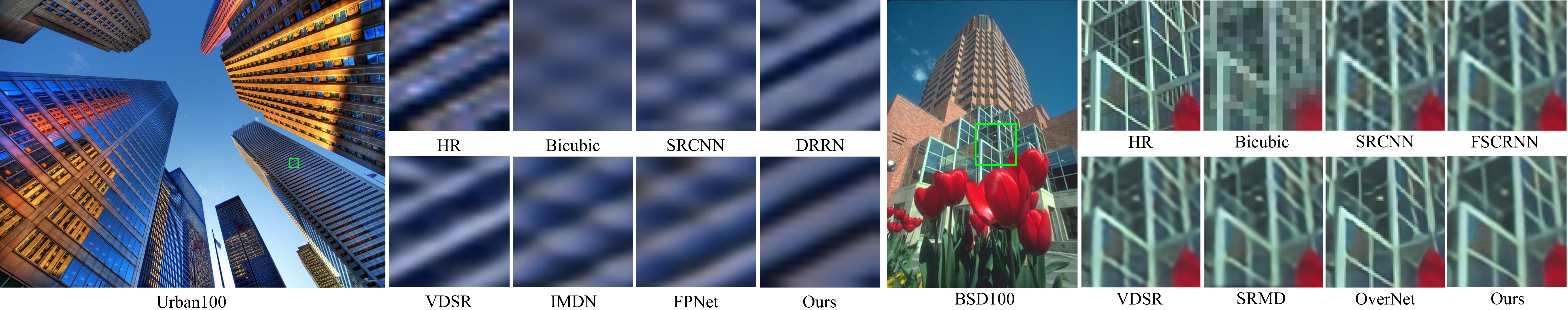}
\end{center}
\label{fig:qualitative_super}
\vspace{-0.35cm}
  \caption{Comparison on the Urban100 and BSD100 dataset.} \vspace{-0.35cm}
\end{figure*}

For KITTI, we aim at the target compactness of $P=1.45M$ with $\alpha=0.55$.  We train our candidate architectures with the same self-supervised procedure proposed by~\cite{godard2019digging} and adopted by the state-of-the-art approaches~\cite{aleotti2021real,cipolletta2021energy,poggi2018towards,wofk2019fastdepth}. After the search, we pick the best architecture (LDP-Depth-K) to compare with the baselines and report the performance figures in Table~\ref{tab:eval_kitti}. The LDP-Depth-K model yields competitive results with the baselines while also being the smallest model. We observe that our proposed method provides noticeable improvement from PyD-Net and EQPyD-Net. Examples from Figure~\ref{fig:qualitative_kitti} show that the predicted depth maps from LDP-Depth-K are more accurate and contain fewer artifacts.

\subsection{Semantic Segmentation}
We then deploy LDP for dense classification tasks such as semantic segmentation that aims to predict discrete labels of image pixels. We employ the same backbone structure as in monocular depth estimation experiments and utilize the cross-entropy loss. For this problem, we evaluate our method on the Cityscapes~\cite{cordts2016cityscapes} and COCO-stuff~\cite{caesar2018coco} datasets. Cityscapes is an outdoor dataset containing images of various urban scenarios. The dataset consists of 5K high-quality annotated frames that 19 classes are used for semantic segmentation. Following the standard procedure, we used 2975, 500, and 1525 images for training, validation, and testing, respectively. COCO-stuff was created by annotating dense stuffs (e.g., sky, ground, walls) from the COCO dataset. This dataset can be utilized for image understanding as it contains 91 more stuffs classes compared to the original dataset. For fair comparison, we also use the COCO-Stuff-10K with 9K and 1K for training and testing purposes. We utilize both the mean Intersection-over-Union (MIoU) as well as the pixel accuracy (pixAcc) to assess the performance of our models.

To perform searching on Cityscapes dataset, we set the target compactness $P=1.25M$ with the balance coefficient $\alpha=0.6$ using input image resolution of $512 \times 1024$ for all experiments. We then compare the best-performing model (LDP-Seg-Ci) with recent approaches~\cite{yu2018bisenet,li2019dfanet,orsic2019defense,yu2021lite,chen2019fasterseg,howard2019searching}. Results in Table~\ref{tab:eval_cityscapes} suggest that LDP-Seg-Ci performs on par with state-of-the-art while using fewer parameters than most methods. Although operating at a lower resolution, our generated model outperforms FasterSeg~\cite{chen2019fasterseg} while being $61\%$ more compact in terms of the number of parameters. Moreover, LDP-Seg-Ci also shows clear improvements compared to the MobileNetV3~\cite{howard2019searching} model. Qualitative results in Figure~\ref{fig:qualitative_cityscapes} also show that our model tends to produce more clean with sharper object boundaries and less cluttering than state-of-the-art approaches.

In the case of the COCO-stuff, we aim at the target compactness of $P=4.0M$ with the balance coefficient $\alpha$ set to $0.6$. During searching and testing, we crop the input into $640 \times 640$ resolution. We evaluate the performance of our best architecture (LDP-Seg-CO) with current state-of-the-art methods~\cite{yu2018bisenet,zhao2018icnet,yu2021bisenet} and report the results in Table~\ref{tab:eval_coco}. Our method also achieves good performance for semantic segmentation on the COCO-stuff dataset while using much fewer parameters than competing approaches.

\subsection{Image Super-resolution}
To further assess the applicability of our proposed framework for dense prediction problems, we apply LDP framework for the image super-resolution task. We also employ a similar scheme as in previous experiments with added upsample blocks between decoder and refinement blocks to increase spatial dimension within the network scale. We then perform architecture search and training on the DIV2K~\cite{agustsson2017ntire} dataset. DIV2K is a high-quality image super-resolution dataset consisting of 800 samples for training, 100 for validation, and 100 for testing purposes. After that, we test our generated models on standard benchmarks such as Set5~\cite{bevilacqua2012low}, Set14~\cite{zeyde2010single}, B100~\cite{arbelaez2010contour}, and Urban100~\cite{huang2015single}. The results of our method are evaluated using the peak signal-to-noise ratio (PSNR) and structural similarity index (SSIM) metrics on the Y channel of the YCbCr color space.

\begin{table}[!b]
\vspace{-0.45cm}
\caption{\label{tab:runtime_comparison}Average runtime comparison of the proposed method and other lightweight models. Runtime values are measured using a Pixel 3a phone with input image resolution ($640 \times 480$).} 
\centering
\small
\begin{tabular}{lc}
\hline
\textbf{Architecture} & \textbf{CPU(ms)} \\ \hline

Ef+FBNet~\cite{tu2020efficient,wu2019fbnet} & 852   \\ \hline

FSRCNN~\cite{dong2016accelerating} & 789  \\ \hline

FastDepth~\cite{wofk2019fastdepth} & 458  \\ \hline

VDSR~\cite{kim2016accurate} & 425 \\ \hline

PyD-Net~\cite{poggi2018towards} & 226  \\ \hline

Lite-HRNet~\cite{yu2021lite} & 217  \\ \hline

LiDNAS-K~\cite{huynh2021lightweight} & 205  \\ \hline

LDP-Seg-Ci & 207 \\ \hline

LDP-Depth-K & 204 \\ \hline

LDP-Sup-x2 & \textbf{203}  \\ \hline

\end{tabular}
\end{table}

\begin{figure*}[!t]
\begin{center}
  \includegraphics[width=0.9\linewidth]{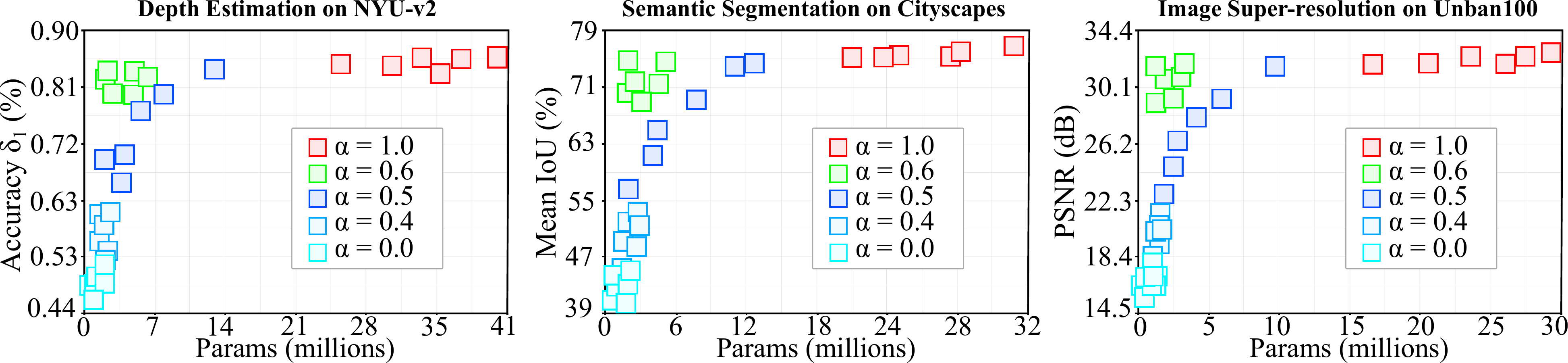}
\end{center}
\vspace{-0.35cm}
  \caption{Trade-off between accuracy vs. the number of parameters of best models trained with different searching scenarios on NYU-Depth-v2, Cityscapes and testing on Urban100 dataset.} \vspace{-0.45cm}
\label{fig:searching_scenarios} %
\end{figure*}

We search for optimized models at super-resolution scales $\times 2$ and $\times 4$ on the DIV2K dataset. In both cases, we determine the target compactness of $P=0.8$ with the balance coefficient $\alpha = 0.55$. We then compare the best-performing architectures (LDP-Sup-x2 and LDP-Sup-x4) with recent methods~\cite{esmaeilzehi2021fpnet,ahn2021neural,ahn2018fast,hui2019lightweight,chu2021fast,lai2017deep} on various testing benchmarks. Tables 10 and 11 show that our models produce more competitive results than several state-of-the-art image super-resolution methods while being relatively compact. Our models perform on par with FPNet while being at least $32\%$ smaller in terms of network size. Figure 10 provides visual comparisons on BSD100 and Urban100 benchmarks. The proposed method yields more accurate results than VDSR~\cite{kim2016accurate}, DRRN~\cite{tai2017image}, FSCRNN~\cite{dong2016accelerating} and more precise details than SRCNN~\cite{dong2014learning}, IMDN~\cite{hui2019lightweight}, FPNet~\cite{esmaeilzehi2021fpnet} methods.

\begin{figure}[!b]
\begin{center}
  \includegraphics[width=0.72\linewidth]{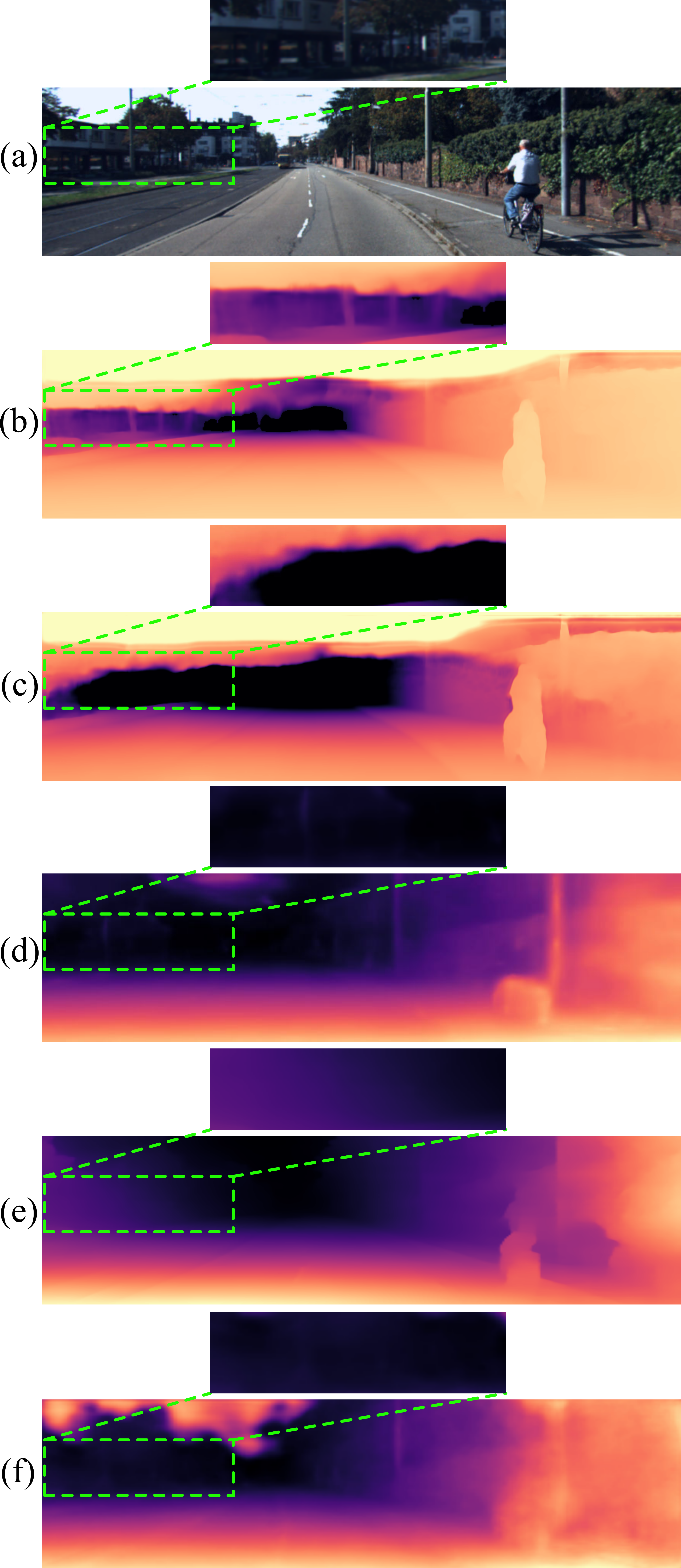}
\end{center}
\vspace{-0.35cm}
  \caption{Comparison on the Eigen split of KITTI. 
  (a) input image,
  (b) LDP-Depth-K,
  (c) LiDNAS-K~\cite{huynh2021lightweight},
  (d) DSNet~\cite{aleotti2021real},
  (e) PyD-Net~\cite{poggi2018towards},
  and (f) FastDepth~\cite{wofk2019fastdepth}.
  Images in the right column presented zoom-in view for better visualization.}
\label{fig:qualitative_kitti} \vspace{-0.35cm}
\end{figure}

\begin{figure*}[!t]
\begin{center}
  \includegraphics[width=0.84\linewidth]{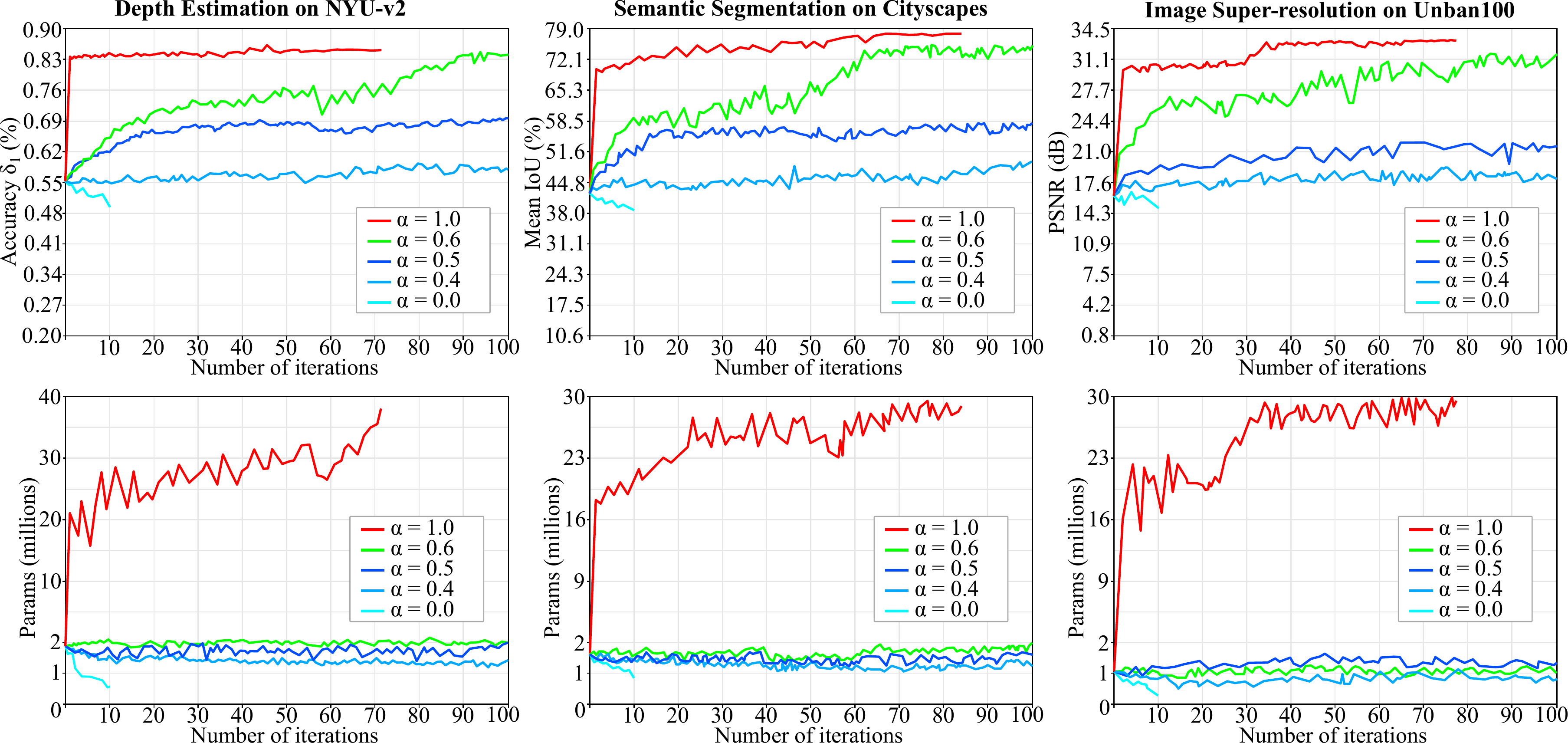}
\end{center}
\vspace{-0.35cm}
  \caption{The progress of different searching scenarios on the NYU-Depth-v2, Cityscapes and testing on Urban100 dataset. Charts show the metrics including thresholded accuracy, Mean IoU, peak-signal-to-noise-ratio (PSNR) and the number of parameters vs. the number of searching iterations.} \vspace{-0.25cm}
\label{fig:convergence_v2}
\end{figure*}

\begin{figure}[!b]
\begin{center}
  \includegraphics[width=0.99\linewidth]{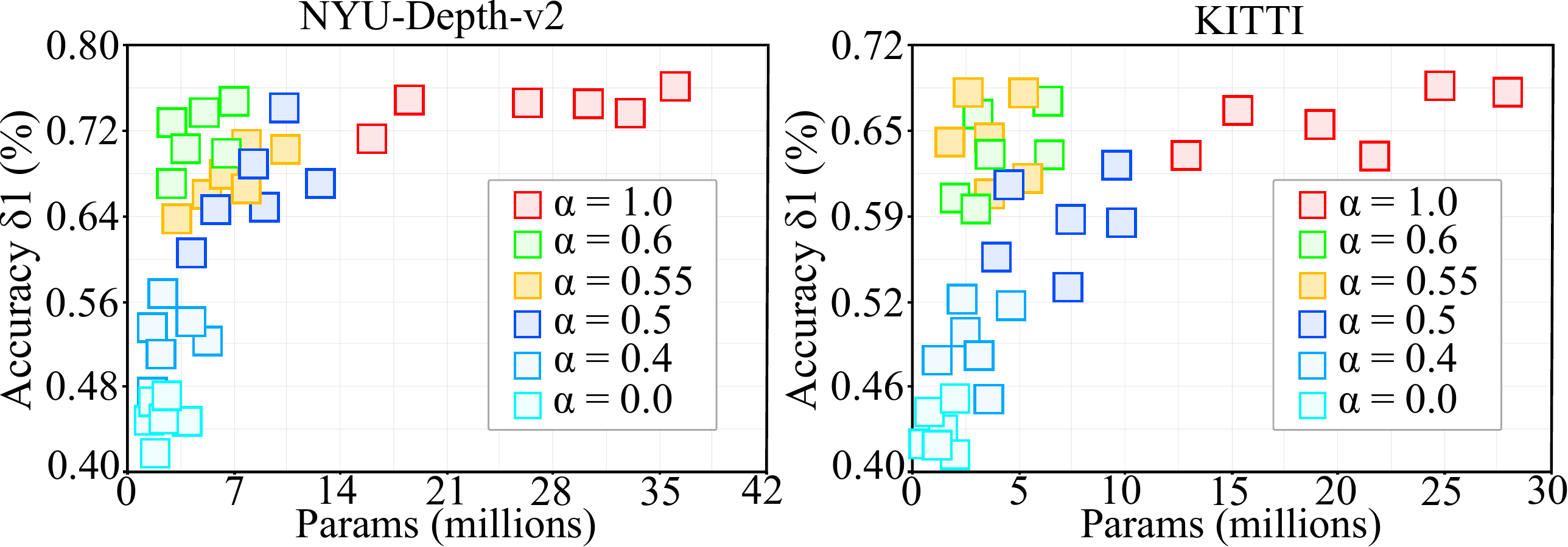}
\end{center}
\vspace{-0.25cm}
  \caption{Grid search using randomly subsampled sets from the training data to look for good balance coefficient values on NYU-Depth-v2 (left) and KITTI (right).}
\label{fig:grid_search} \vspace{-0.25cm}
\end{figure}

\subsection{Runtime Measurement}
We also compare the runtime of our models with state-of-the-art lightweight methods on an Android device using the app from the Mobile AI benchmark developed by Ignatov et al.~\cite{ignatov2021fast}. To this end, we utilize the pre-trained models provided by the authors (Tensorflow~\cite{poggi2018towards}, PyTorch~\cite{wofk2019fastdepth}), convert them to \textit{tflite} and measure their runtime on mobile CPUs. 
The results in Table~\ref{tab:runtime_comparison} suggest that the proposed approaches produce competing performance, with the potential of running real-time on mobile devices with further optimization.

\subsection{Exploration Convergence}

We experiment with various settings for the multi-objective balance coefficient ($\alpha$) to assess its effect on the performance. For this purpose, we perform the architecture search with $\alpha$ set to $0.0$, $0.4$, $0.5$, $0.6$, and $1.0$ while the target compactness $P=2.0M$.  Figure~\ref{fig:convergence_v2} presents the searching progress for accuracy (left), the number of parameters (center), and validation grade (right) from one parent architecture on NYU-Depth-v2. We observe that, scenario with $\alpha=0.0$ quickly becomes saturated as it only gives reward to the smallest model. Searching with $\alpha=0.4$ favors models with compact size but also with limited accuracy. The case with $\alpha=0.5$ provides a more balanced option, but accuracy is hindered due to fluctuation during searching. The exploration with $\alpha=1.0$ seeks for the network with the best accuracy yet producing significantly larger architecture while the case where $\alpha=0.6$ achieves promising accuracy although with slightly bigger model than the target compactness.

\begin{figure*}[ht]
\begin{center}
  \includegraphics[width=0.935\linewidth]{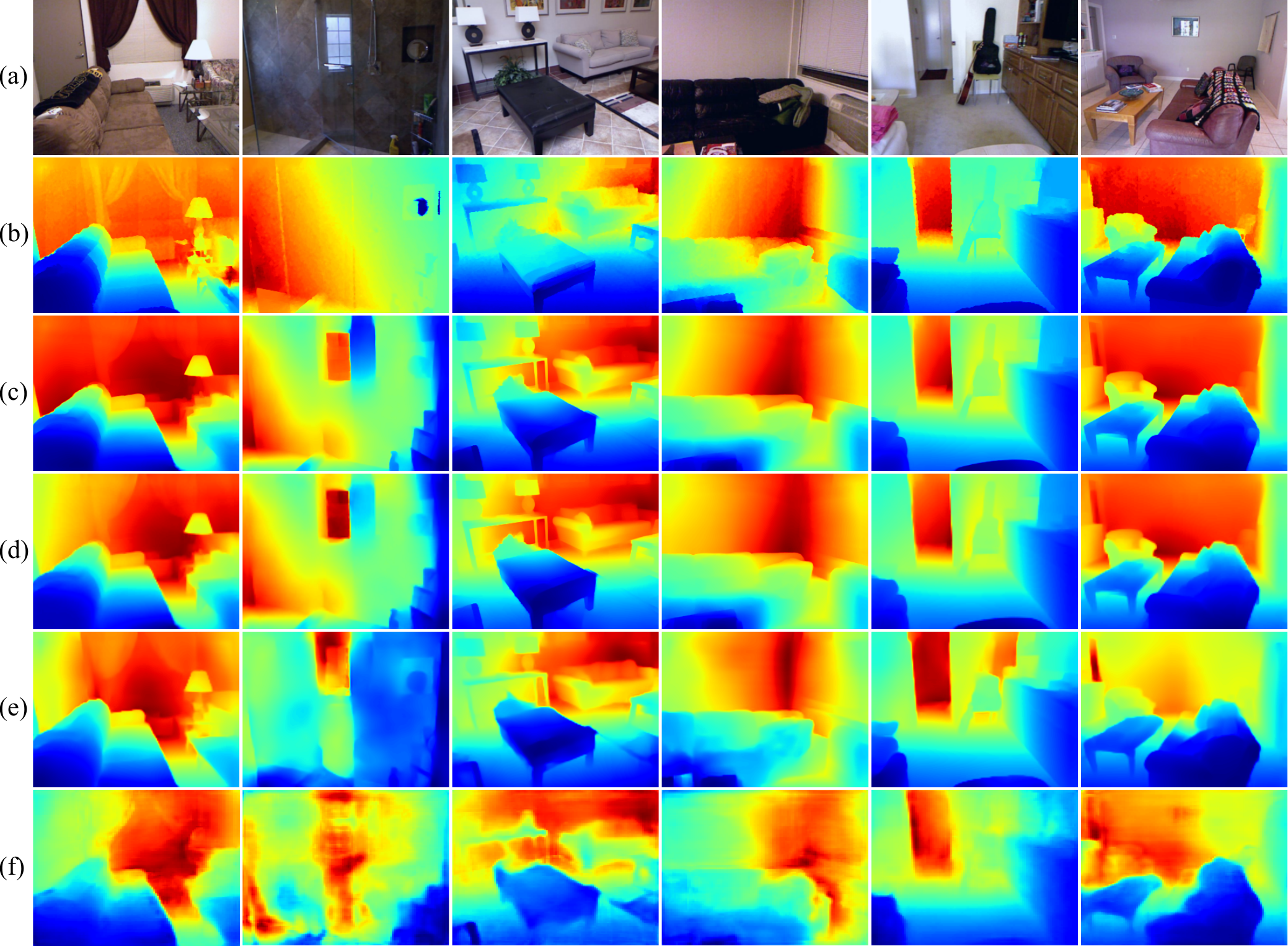}
\end{center}
\vspace{-0.5cm}
  \caption{Comparison on the NYU test set. 
  (a) input image,
  (b) ground truth,
  (c) LDP-Depth-N,
  (d) LiDNAS-N~\cite{huynh2021lightweight},
  (e) Ef+FBNet~\cite{tu2020efficient,wu2019fbnet},
  and (f) FastDepth~\cite{wofk2019fastdepth}.}
\label{fig:qualitative_nyu} \vspace{-0.5cm}
\end{figure*}

\subsection{Searching Scenarios}
To further analyze the outcome of different searching scenarios, we perform architecture searches for \textit{six} parent networks in five settings with $\alpha=0.0, 0.4, 0.5, 0.6, 1.0$ and $P=2.0M$ on NYU-Depth-v2. Results in Figure~\ref{fig:searching_scenarios} show that the best performance models in the case of $\alpha=0.5$ are more spread out, while training instances with $\alpha=0.6$ tend to produce both accurate and lightweight architectures. This, in turn, emphasizes the trade-off between validation accuracy and the network size.

\subsection{Balance Coefficient Search}
Determining a good balance coefficient value $(\alpha)$ for Eq.~\ref{eq:validation_grade} and~\ref{eq:mutation_exploration_reward} is crucial as it greatly affects the search performance. To this end, we perform grid search on randomly subsampled sets from the training data seeking for the optimized $\alpha$ value. The pattern in Figure~\ref{fig:grid_search} shows that, for NYU-Depth-v2 [4] and KITTI [2], approximately good $\alpha$ values range from $0.5$ to $0.6$. Additionally, the grid search is much faster (only requires $\sim 15$ hours on one dataset), enabling finding good $\alpha$ values when deploying to different datasets.

\section{Conclusion}

\noindent This paper proposed a novel NAS framework to construct lightweight dense prediction architectures using Assisted Tabu Search and employing a well-defined search space for balancing layer diversity and search volume. The proposed method achieves competitive accuracy on diverse datasets while running faster on mobile devices and being more compact than state-of-the-art handcrafted and automatically generated models. Our work provides a potential approach towards optimizing the accuracy and the network size for dense prediction without the need for manual tweaking of deep neural architectures.

{\small
\bibliographystyle{ieee_fullname}
\bibliography{main}

\begin{thebibliography}{100}\itemsep=-1pt

\bibitem{agustsson2017ntire}
Eirikur Agustsson and Radu Timofte.
\newblock Ntire 2017 challenge on single image super-resolution: Dataset and
  study.
\newblock In {\em CVPR workshops}, pages 126--135, 2017.

\bibitem{ahn2021neural}
Joon~Young Ahn and Nam~Ik Cho.
\newblock Neural architecture search for image super-resolution using densely
  constructed search space: Deconas.
\newblock In {\em 2020 25th International Conference on Pattern Recognition
  (ICPR)}, pages 4829--4836. IEEE, 2021.

\bibitem{ahn2018fast}
Namhyuk Ahn, Byungkon Kang, and Kyung-Ah Sohn.
\newblock Fast, accurate, and lightweight super-resolution with cascading
  residual network.
\newblock In {\em Proceedings of the European Conference on Computer Vision
  (ECCV)}, pages 252--268, 2018.

\bibitem{aleotti2021real}
Filippo Aleotti, Giulio Zaccaroni, Luca Bartolomei, Matteo Poggi, Fabio Tosi,
  and Stefano Mattoccia.
\newblock Real-time single image depth perception in the wild with handheld
  devices.
\newblock {\em Sensors}, 21(1):15, 2021.

\bibitem{arbelaez2010contour}
Pablo Arbelaez, Michael Maire, Charless Fowlkes, and Jitendra Malik.
\newblock Contour detection and hierarchical image segmentation.
\newblock {\em IEEE transactions on pattern analysis and machine intelligence},
  33(5):898--916, 2010.

\bibitem{badrinarayanan2017segnet}
Vijay Badrinarayanan, Alex Kendall, and Roberto Cipolla.
\newblock Segnet: A deep convolutional encoder-decoder architecture for image
  segmentation.
\newblock {\em IEEE transactions on pattern analysis and machine intelligence},
  39(12):2481--2495, 2017.

\bibitem{baker2016designing}
Bowen Baker, Otkrist Gupta, Nikhil Naik, and Ramesh Raskar.
\newblock Designing neural network architectures using reinforcement learning.
\newblock {\em arXiv preprint arXiv:1611.02167}, 2016.

\bibitem{behjati2021overnet}
Parichehr Behjati, Pau Rodriguez, Armin Mehri, Isabelle Hupont,
  Carles~Fernandez Tena, and Jordi Gonzalez.
\newblock Overnet: Lightweight multi-scale super-resolution with overscaling
  network.
\newblock In {\em Proceedings of the IEEE/CVF Winter Conference on Applications
  of Computer Vision}, pages 2694--2703, 2021.

\bibitem{bevilacqua2012low}
Marco Bevilacqua, Aline Roumy, Christine Guillemot, and Marie~Line
  Alberi-Morel.
\newblock Low-complexity single-image super-resolution based on nonnegative
  neighbor embedding.
\newblock {\em British Machine Vision Conference (BMVC)}, 2012.

\bibitem{bhat2021adabins}
Shariq~Farooq Bhat, Ibraheem Alhashim, and Peter Wonka.
\newblock Adabins: Depth estimation using adaptive bins.
\newblock In {\em Proceedings of the IEEE conference on computer vision and
  pattern recognition}, pages 4009--4018, 2021.

\bibitem{caesar2018coco}
Holger Caesar, Jasper Uijlings, and Vittorio Ferrari.
\newblock Coco-stuff: Thing and stuff classes in context.
\newblock In {\em CVPR}, pages 1209--1218, 2018.

\bibitem{chen2017deeplab}
Liang-Chieh Chen, George Papandreou, Iasonas Kokkinos, Kevin Murphy, and Alan~L
  Yuille.
\newblock Deeplab: Semantic image segmentation with deep convolutional nets,
  atrous convolution, and fully connected crfs.
\newblock {\em IEEE transactions on pattern analysis and machine intelligence},
  40(4):834--848, 2017.

\bibitem{chen2017rethinking}
Liang-Chieh Chen, George Papandreou, Florian Schroff, and Hartwig Adam.
\newblock Rethinking atrous convolution for semantic image segmentation.
\newblock {\em arXiv preprint arXiv:1706.05587}, 2017.

\bibitem{chen2018encoder}
Liang-Chieh Chen, Yukun Zhu, George Papandreou, Florian Schroff, and Hartwig
  Adam.
\newblock Encoder-decoder with atrous separable convolution for semantic image
  segmentation.
\newblock In {\em Proceedings of the European conference on computer vision
  (ECCV)}, pages 801--818, 2018.

\bibitem{chen2016single}
Weifeng Chen, Zhao Fu, Dawei Yang, and Jia Deng.
\newblock Single-image depth perception in the wild.
\newblock {\em Advances in neural information processing systems}, 29:730--738,
  2016.

\bibitem{chen2019fasterseg}
Wuyang Chen, Xinyu Gong, Xianming Liu, Qian Zhang, Yuan Li, and Zhangyang Wang.
\newblock Fasterseg: Searching for faster real-time semantic segmentation.
\newblock {\em arXiv preprint arXiv:1912.10917}, 2019.

\bibitem{chen2019structure}
Xiaotian Chen, Xuejin Chen, and Zheng-Jun Zha.
\newblock Structure-aware residual pyramid network for monocular depth
  estimation.
\newblock In {\em Proceedings of the 28th International Joint Conference on
  Artificial Intelligence}, pages 694--700. AAAI Press, 2019.

\bibitem{chu2021fast}
Xiangxiang Chu, Bo Zhang, Hailong Ma, Ruijun Xu, and Qingyuan Li.
\newblock Fast, accurate and lightweight super-resolution with neural
  architecture search.
\newblock In {\em 2020 25th International Conference on Pattern Recognition
  (ICPR)}, pages 59--64. IEEE, 2021.

\bibitem{cipolletta2021energy}
Antonio Cipolletta, Valentino Peluso, Andrea Calimera, Matteo Poggi, Fabio
  Tosi, Filippo Aleotti, and Stefano Mattoccia.
\newblock Energy-quality scalable monocular depth estimation on low-power cpus.
\newblock {\em IEEE Internet of Things Journal}, 2021.

\bibitem{cordts2016cityscapes}
Marius Cordts, Mohamed Omran, Sebastian Ramos, Timo Rehfeld, Markus Enzweiler,
  Rodrigo Benenson, Uwe Franke, Stefan Roth, and Bernt Schiele.
\newblock The cityscapes dataset for semantic urban scene understanding.
\newblock In {\em Proceedings of the IEEE conference on computer vision and
  pattern recognition}, pages 3213--3223, 2016.

\bibitem{dai2015boxsup}
Jifeng Dai, Kaiming He, and Jian Sun.
\newblock Boxsup: Exploiting bounding boxes to supervise convolutional networks
  for semantic segmentation.
\newblock In {\em Proceedings of the IEEE international conference on computer
  vision}, pages 1635--1643, 2015.

\bibitem{dong2014learning}
Chao Dong, Chen~Change Loy, Kaiming He, and Xiaoou Tang.
\newblock Learning a deep convolutional network for image super-resolution.
\newblock In {\em European conference on computer vision}, pages 184--199.
  Springer, 2014.

\bibitem{dong2015image}
Chao Dong, Chen~Change Loy, Kaiming He, and Xiaoou Tang.
\newblock Image super-resolution using deep convolutional networks.
\newblock {\em IEEE transactions on pattern analysis and machine intelligence},
  38(2):295--307, 2015.

\bibitem{dong2016accelerating}
Chao Dong, Chen~Change Loy, and Xiaoou Tang.
\newblock Accelerating the super-resolution convolutional neural network.
\newblock In {\em European conference on computer vision}, pages 391--407.
  Springer, 2016.

\bibitem{dong2018dpp}
Jin-Dong Dong, An-Chieh Cheng, Da-Cheng Juan, Wei Wei, and Min Sun.
\newblock Dpp-net: Device-aware progressive search for pareto-optimal neural
  architectures.
\newblock In {\em Proceedings of the European Conference on Computer Vision
  (ECCV)}, pages 517--531, 2018.

\bibitem{eigen2015predicting}
David Eigen and Rob Fergus.
\newblock Predicting depth, surface normals and semantic labels with a common
  multi-scale convolutional architecture.
\newblock In {\em Proceedings of the IEEE International Conference on Computer
  Vision}, pages 2650--2658, 2015.

\bibitem{eigen2014depth}
David Eigen, Christian Puhrsch, and Rob Fergus.
\newblock Depth map prediction from a single image using a multi-scale deep
  network.
\newblock In {\em Advances in neural information processing systems}, pages
  2366--2374, 2014.

\bibitem{elsken2018efficient}
Thomas Elsken, Jan~Hendrik Metzen, and Frank Hutter.
\newblock Efficient multi-objective neural architecture search via lamarckian
  evolution.
\newblock {\em arXiv preprint arXiv:1804.09081}, 2018.

\bibitem{elsken2018multi}
Thomas Elsken, Jan~Hendrik Metzen, and Frank Hutter.
\newblock Multi-objective architecture search for cnns.
\newblock {\em arXiv preprint arXiv:1804.09081}, 2, 2018.

\bibitem{esmaeilzehi2021fpnet}
Alireza Esmaeilzehi, M~Omair Ahmad, and MNS Swamy.
\newblock Fpnet: A deep light-weight interpretable neural network using forward
  prediction filtering for efficient single image super resolution.
\newblock {\em IEEE Transactions on Circuits and Systems II: Express Briefs},
  2021.

\bibitem{facil2019cam}
Jose~M Facil, Benjamin Ummenhofer, Huizhong Zhou, Luis Montesano, Thomas Brox,
  and Javier Civera.
\newblock Cam-convs: camera-aware multi-scale convolutions for single-view
  depth.
\newblock In {\em IEEE/CVF Conference on Computer Vision and Pattern
  Recognition}, pages 11826--11835, 2019.

\bibitem{fu2018deep}
Huan Fu, Mingming Gong, Chaohui Wang, Kayhan Batmanghelich, and Dacheng Tao.
\newblock Deep ordinal regression network for monocular depth estimation.
\newblock In {\em CVPR}, pages 2002--2011, 2018.

\bibitem{garg2016unsupervised}
Ravi Garg, Vijay~Kumar Bg, Gustavo Carneiro, and Ian Reid.
\newblock Unsupervised cnn for single view depth estimation: Geometry to the
  rescue.
\newblock In {\em European conference on computer vision}, pages 740--756.
  Springer, 2016.

\bibitem{geiger2013vision}
Andreas Geiger, Philip Lenz, Christoph Stiller, and Raquel Urtasun.
\newblock Vision meets robotics: The kitti dataset.
\newblock {\em The International Journal of Robotics Research},
  32(11):1231--1237, 2013.

\bibitem{girshick2014rich}
Ross Girshick, Jeff Donahue, Trevor Darrell, and Jitendra Malik.
\newblock Rich feature hierarchies for accurate object detection and semantic
  segmentation.
\newblock In {\em Proceedings of the IEEE conference on computer vision and
  pattern recognition}, pages 580--587, 2014.

\bibitem{glover1986future}
Fred Glover.
\newblock Future paths for integer programming and links to artificial
  intelligence.
\newblock {\em Computers \& operations research}, 13(5):533--549, 1986.

\bibitem{godard2019digging}
Cl{\'e}ment Godard, Oisin Mac~Aodha, Michael Firman, and Gabriel~J Brostow.
\newblock Digging into self-supervised monocular depth estimation.
\newblock In {\em Proceedings of the IEEE conference on computer vision and
  pattern recognition}, pages 3828--3838, 2019.

\bibitem{gonzalezbello2020forget}
Juan~Luis GonzalezBello and Munchurl Kim.
\newblock Forget about the lidar: Self-supervised depth estimators with med
  probability volumes.
\newblock {\em Advances in Neural Information Processing Systems}, 33, 2020.

\bibitem{han2020ghostnet}
Kai Han, Yunhe Wang, Qi Tian, Jianyuan Guo, Chunjing Xu, and Chang Xu.
\newblock Ghostnet: More features from cheap operations.
\newblock In {\em Proceedings of the IEEE conference on computer vision and
  pattern recognition}, pages 1580--1589, 2020.

\bibitem{han2015deep}
Song Han, Huizi Mao, and William~J Dally.
\newblock Deep compression: Compressing deep neural networks with pruning,
  trained quantization and huffman coding.
\newblock {\em arXiv preprint arXiv:1510.00149}, 2015.

\bibitem{hariharan2014simultaneous}
Bharath Hariharan, Pablo Arbel{\'a}ez, Ross Girshick, and Jitendra Malik.
\newblock Simultaneous detection and segmentation.
\newblock In {\em European conference on computer vision}, pages 297--312.
  Springer, 2014.

\bibitem{hariharan2015hypercolumns}
Bharath Hariharan, Pablo Arbel{\'a}ez, Ross Girshick, and Jitendra Malik.
\newblock Hypercolumns for object segmentation and fine-grained localization.
\newblock In {\em Proceedings of the IEEE conference on computer vision and
  pattern recognition}, pages 447--456, 2015.

\bibitem{he2019ode}
Xiangyu He, Zitao Mo, Peisong Wang, Yang Liu, Mingyuan Yang, and Jian Cheng.
\newblock Ode-inspired network design for single image super-resolution.
\newblock In {\em Proceedings of the IEEE conference on computer vision and
  pattern recognition}, pages 1732--1741, 2019.

\bibitem{howard2019searching}
Andrew Howard, Mark Sandler, Grace Chu, Liang-Chieh Chen, Bo Chen, Mingxing
  Tan, Weijun Wang, Yukun Zhu, Ruoming Pang, Vijay Vasudevan, et~al.
\newblock Searching for mobilenetv3.
\newblock In {\em Proceedings of the IEEE/CVF International Conference on
  Computer Vision}, pages 1314--1324, 2019.

\bibitem{howard2017mobilenets}
Andrew~G Howard, Menglong Zhu, Bo Chen, Dmitry Kalenichenko, Weijun Wang,
  Tobias Weyand, Marco Andreetto, and Hartwig Adam.
\newblock Mobilenets: Efficient convolutional neural networks for mobile vision
  applications.
\newblock {\em arXiv preprint arXiv:1704.04861}, 2017.

\bibitem{hsu2018monas}
Chi-Hung Hsu, Shu-Huan Chang, Jhao-Hong Liang, Hsin-Ping Chou, Chun-Hao Liu,
  Shih-Chieh Chang, Jia-Yu Pan, Yu-Ting Chen, Wei Wei, and Da-Cheng Juan.
\newblock Monas: Multi-objective neural architecture search using reinforcement
  learning.
\newblock {\em arXiv preprint arXiv:1806.10332}, 2018.

\bibitem{Hu2018RevisitingSI}
Junjie Hu, Mete Ozay, Yan Zhang, and Takayuki Okatani.
\newblock Revisiting single image depth estimation: Toward higher resolution
  maps with accurate object boundaries.
\newblock In {\em IEEE Winter Conf. on Applications of Computer Vision (WACV)},
  2019.

\bibitem{huang2015single}
Jia-Bin Huang, Abhishek Singh, and Narendra Ahuja.
\newblock Single image super-resolution from transformed self-exemplars.
\newblock In {\em CVPR}, pages 5197--5206, 2015.

\bibitem{hui2019lightweight}
Zheng Hui, Xinbo Gao, Yunchu Yang, and Xiumei Wang.
\newblock Lightweight image super-resolution with information
  multi-distillation network.
\newblock In {\em Proceedings of the 27th ACM International Conference on
  Multimedia}, pages 2024--2032, 2019.

\bibitem{hui2018fast}
Zheng Hui, Xiumei Wang, and Xinbo Gao.
\newblock Fast and accurate single image super-resolution via information
  distillation network.
\newblock In {\em Proceedings of the IEEE conference on computer vision and
  pattern recognition}, pages 723--731, 2018.

\bibitem{huynh2021lightweight}
Lam Huynh, Phong Nguyen, Jiri Matas, Esa Rahtu, and Janne Heikkila.
\newblock Lightweight monocular depth with a novel neural architecture search
  method.
\newblock {\em arXiv preprint arXiv:2108.11105}, 2021.

\bibitem{huynh2020guiding}
Lam Huynh, Phong Nguyen-Ha, Jiri Matas, Esa Rahtu, and Janne Heikkil{\"a}.
\newblock Guiding monocular depth estimation using depth-attention volume.
\newblock In {\em European Conference on Computer Vision}, pages 581--597.
  Springer, 2020.

\bibitem{ignatov2021fast}
Andrey Ignatov, Grigory Malivenko, David Plowman, Samarth Shukla, Radu Timofte,
  Ziyu Zhang, Yicheng Wang, Zilong Huang, Guozhong Luo, Gang Yu, et~al.
\newblock Fast and accurate single-image depth estimation on mobile devices,
  mobile ai 2021 challenge: Report.
\newblock {\em arXiv preprint arXiv:2105.08630}, 2021.

\bibitem{jain2021semask}
Jitesh Jain, Anukriti Singh, Nikita Orlov, Zilong Huang, Jiachen Li, Steven
  Walton, and Humphrey Shi.
\newblock Semask: Semantically masked transformers for semantic segmentation.
\newblock {\em arXiv preprint arXiv:2112.12782}, 2021.

\bibitem{jiao2018look}
Jianbo Jiao, Ying Cao, Yibing Song, and Rynson Lau.
\newblock Look deeper into depth: Monocular depth estimation with semantic
  booster and attention-driven loss.
\newblock In {\em Proceedings of the European Conference on Computer Vision
  (ECCV)}, pages 53--69, 2018.

\bibitem{kim2016accurate}
Jiwon Kim, Jung~Kwon Lee, and Kyoung~Mu Lee.
\newblock Accurate image super-resolution using very deep convolutional
  networks.
\newblock In {\em Proceedings of the IEEE conference on computer vision and
  pattern recognition}, pages 1646--1654, 2016.

\bibitem{kim2016deeply}
Jiwon Kim, Jung~Kwon Lee, and Kyoung~Mu Lee.
\newblock Deeply-recursive convolutional network for image super-resolution.
\newblock In {\em Proceedings of the IEEE conference on computer vision and
  pattern recognition}, pages 1637--1645, 2016.

\bibitem{kingma2014adam}
Diederik~P Kingma and Jimmy Ba.
\newblock Adam: A method for stochastic optimization.
\newblock {\em arXiv preprint arXiv:1412.6980}, 2014.

\bibitem{krahenbuhl2011efficient}
Philipp Kr{\"a}henb{\"u}hl and Vladlen Koltun.
\newblock Efficient inference in fully connected crfs with gaussian edge
  potentials.
\newblock {\em Advances in neural information processing systems}, 24:109--117,
  2011.

\bibitem{lai2017deep}
Wei-Sheng Lai, Jia-Bin Huang, Narendra Ahuja, and Ming-Hsuan Yang.
\newblock Deep laplacian pyramid networks for fast and accurate
  super-resolution.
\newblock In {\em Proceedings of the IEEE conference on computer vision and
  pattern recognition}, pages 624--632, 2017.

\bibitem{laina2016deeper}
Iro Laina, Christian Rupprecht, Vasileios Belagiannis, Federico Tombari, and
  Nassir Navab.
\newblock Deeper depth prediction with fully convolutional residual networks.
\newblock In {\em 2016 Fourth International Conference on 3D Vision (3DV)},
  pages 239--248. IEEE, 2016.

\bibitem{lee2019big}
Jin~Han Lee, Myung-Kyu Han, Dong~Wook Ko, and Il~Hong Suh.
\newblock From big to small: Multi-scale local planar guidance for monocular
  depth estimation.
\newblock {\em arXiv preprint arXiv:1907.10326}, 2019.

\bibitem{lee2019monocular}
Jae-Han Lee and Chang-Su Kim.
\newblock Monocular depth estimation using relative depth maps.
\newblock In {\em CVPR}, pages 9729--9738, 2019.

\bibitem{li2020s}
Biao Li, Bo Wang, Jiabin Liu, Zhiquan Qi, and Yong Shi.
\newblock s-lwsr: Super lightweight super-resolution network.
\newblock {\em IEEE Transactions on Image Processing}, 29:8368--8380, 2020.

\bibitem{li2019dfanet}
Hanchao Li, Pengfei Xiong, Haoqiang Fan, and Jian Sun.
\newblock Dfanet: Deep feature aggregation for real-time semantic segmentation.
\newblock In {\em Proceedings of the IEEE conference on computer vision and
  pattern recognition}, pages 9522--9531, 2019.

\bibitem{li2021micronet}
Yunsheng Li, Yinpeng Chen, Xiyang Dai, Dongdong Chen, Mengchen Liu, Lu Yuan,
  Zicheng Liu, Lei Zhang, and Nuno Vasconcelos.
\newblock Micronet: Improving image recognition with extremely low flops.
\newblock In {\em Proceedings of the IEEE/CVF International Conference on
  Computer Vision}, pages 468--477, 2021.

\bibitem{li2019feedback}
Zhen Li, Jinglei Yang, Zheng Liu, Xiaomin Yang, Gwanggil Jeon, and Wei Wu.
\newblock Feedback network for image super-resolution.
\newblock In {\em Proceedings of the IEEE conference on computer vision and
  pattern recognition}, pages 3867--3876, 2019.

\bibitem{liang2015semantic}
Chen Liang-Chieh, George Papandreou, Iasonas Kokkinos, Kevin Murphy, and Alan
  Yuille.
\newblock Semantic image segmentation with deep convolutional nets and fully
  connected crfs.
\newblock In {\em International Conference on Learning Representations}, 2015.

\bibitem{liu2019planercnn}
Chen Liu, Kihwan Kim, Jinwei Gu, Yasutaka Furukawa, and Jan Kautz.
\newblock Planercnn: 3d plane detection and reconstruction from a single image.
\newblock In {\em CVPR}, pages 4450--4459, 2019.

\bibitem{liu2018planenet}
Chen Liu, Jimei Yang, Duygu Ceylan, Ersin Yumer, and Yasutaka Furukawa.
\newblock Planenet: Piece-wise planar reconstruction from a single rgb image.
\newblock In {\em CVPR}, pages 2579--2588, 2018.

\bibitem{liu2018progressive}
Chenxi Liu, Barret Zoph, Maxim Neumann, Jonathon Shlens, Wei Hua, Li-Jia Li, Li
  Fei-Fei, Alan Yuille, Jonathan Huang, and Kevin Murphy.
\newblock Progressive neural architecture search.
\newblock In {\em Proceedings of the European conference on computer vision
  (ECCV)}, pages 19--34, 2018.

\bibitem{liu2018darts}
Hanxiao Liu, Karen Simonyan, and Yiming Yang.
\newblock Darts: Differentiable architecture search.
\newblock {\em arXiv preprint arXiv:1806.09055}, 2018.

\bibitem{liu2021swin}
Ze Liu, Han Hu, Yutong Lin, Zhuliang Yao, Zhenda Xie, Yixuan Wei, Jia Ning, Yue
  Cao, Zheng Zhang, Li Dong, et~al.
\newblock Swin transformer v2: Scaling up capacity and resolution.
\newblock {\em arXiv preprint arXiv:2111.09883}, 2021.

\bibitem{long2015fully}
Jonathan Long, Evan Shelhamer, and Trevor Darrell.
\newblock Fully convolutional networks for semantic segmentation.
\newblock In {\em Proceedings of the IEEE conference on computer vision and
  pattern recognition}, pages 3431--3440, 2015.

\bibitem{luo2018neural}
Renqian Luo, Fei Tian, Tao Qin, Enhong Chen, and Tie-Yan Liu.
\newblock Neural architecture optimization.
\newblock {\em arXiv preprint arXiv:1808.07233}, 2018.

\bibitem{mellor2021neural}
Joe Mellor, Jack Turner, Amos Storkey, and Elliot~J Crowley.
\newblock Neural architecture search without training.
\newblock In {\em International Conference on Machine Learning}, pages
  7588--7598. PMLR, 2021.

\bibitem{noh2015learning}
Hyeonwoo Noh, Seunghoon Hong, and Bohyung Han.
\newblock Learning deconvolution network for semantic segmentation.
\newblock In {\em Proceedings of the IEEE international conference on computer
  vision}, pages 1520--1528, 2015.

\bibitem{orsic2019defense}
Marin Orsic, Ivan Kreso, Petra Bevandic, and Sinisa Segvic.
\newblock In defense of pre-trained imagenet architectures for real-time
  semantic segmentation of road-driving images.
\newblock In {\em Proceedings of the IEEE conference on computer vision and
  pattern recognition}, pages 12607--12616, 2019.

\bibitem{pham2018efficient}
Hieu Pham, Melody Guan, Barret Zoph, Quoc Le, and Jeff Dean.
\newblock Efficient neural architecture search via parameters sharing.
\newblock In {\em International Conference on Machine Learning}, pages
  4095--4104. PMLR, 2018.

\bibitem{poggi2018towards}
Matteo Poggi, Filippo Aleotti, Fabio Tosi, and Stefano Mattoccia.
\newblock Towards real-time unsupervised monocular depth estimation on cpu.
\newblock In {\em 2018 IEEE/RSJ International Conference on Intelligent Robots
  and Systems (IROS)}, pages 5848--5854. IEEE, 2018.

\bibitem{qi2018geonet}
Xiaojuan Qi, Renjie Liao, Zhengzhe Liu, Raquel Urtasun, and Jiaya Jia.
\newblock Geonet: Geometric neural network for joint depth and surface normal
  estimation.
\newblock In {\em CVPR}, pages 283--291, 2018.

\bibitem{ramamonjisoa2019sharpnet}
Michael Ramamonjisoa and Vincent Lepetit.
\newblock Sharpnet: Fast and accurate recovery of occluding contours in
  monocular depth estimation.
\newblock {\em (ICCV) Workshops}, 2019.

\bibitem{Ranftl2021}
Ren\'{e} Ranftl, Alexey Bochkovskiy, and Vladlen Koltun.
\newblock Vision transformers for dense prediction.
\newblock {\em ArXiv preprint}, 2021.

\bibitem{real2019regularized}
Esteban Real, Alok Aggarwal, Yanping Huang, and Quoc~V Le.
\newblock Regularized evolution for image classifier architecture search.
\newblock In {\em Proceedings of the aaai conference on artificial
  intelligence}, volume~33, pages 4780--4789, 2019.

\bibitem{ronneberger2015u}
Olaf Ronneberger, Philipp Fischer, and Thomas Brox.
\newblock U-net: Convolutional networks for biomedical image segmentation.
\newblock In {\em International Conference on Medical image computing and
  computer-assisted intervention}, pages 234--241. Springer, 2015.

\bibitem{saikia2019autodispnet}
Tonmoy Saikia, Yassine Marrakchi, Arber Zela, Frank Hutter, and Thomas Brox.
\newblock Autodispnet: Improving disparity estimation with automl.
\newblock In {\em Proceedings of the IEEE/CVF International Conference on
  Computer Vision}, pages 1812--1823, 2019.

\bibitem{sandler2018mobilenetv2}
Mark Sandler, Andrew Howard, Menglong Zhu, Andrey Zhmoginov, and Liang-Chieh
  Chen.
\newblock Mobilenetv2: Inverted residuals and linear bottlenecks.
\newblock In {\em Proceedings of the IEEE conference on computer vision and
  pattern recognition}, pages 4510--4520, 2018.

\bibitem{saxena2006learning}
Ashutosh Saxena, Sung~H Chung, and Andrew~Y Ng.
\newblock Learning depth from single monocular images.
\newblock In {\em Advances in neural information processing systems}, pages
  1161--1168, 2006.

\bibitem{silberman2012indoor}
Nathan Silberman, Derek Hoiem, Pushmeet Kohli, and Rob Fergus.
\newblock Indoor segmentation and support inference from rgbd images.
\newblock In {\em European Conference on Computer Vision}, pages 746--760.
  Springer, 2012.

\bibitem{sun2018pwc}
Deqing Sun, Xiaodong Yang, Ming-Yu Liu, and Jan Kautz.
\newblock Pwc-net: Cnns for optical flow using pyramid, warping, and cost
  volume.
\newblock In {\em CVPR}, pages 8934--8943, 2018.

\bibitem{tai2017image}
Ying Tai, Jian Yang, and Xiaoming Liu.
\newblock Image super-resolution via deep recursive residual network.
\newblock In {\em Proceedings of the IEEE conference on computer vision and
  pattern recognition}, pages 3147--3155, 2017.

\bibitem{tai2017memnet}
Ying Tai, Jian Yang, Xiaoming Liu, and Chunyan Xu.
\newblock Memnet: A persistent memory network for image restoration.
\newblock In {\em Proceedings of the IEEE international conference on computer
  vision}, pages 4539--4547, 2017.

\bibitem{tan2019mnasnet}
Mingxing Tan, Bo Chen, Ruoming Pang, Vijay Vasudevan, Mark Sandler, Andrew
  Howard, and Quoc~V Le.
\newblock Mnasnet: Platform-aware neural architecture search for mobile.
\newblock In {\em CVPR}, pages 2820--2828, 2019.

\bibitem{tu2020efficient}
Xiaohan Tu, Cheng Xu, Siping Liu, Renfa Li, Guoqi Xie, Jing Huang, and
  Laurence~Tianruo Yang.
\newblock Efficient monocular depth estimation for edge devices in internet of
  things.
\newblock {\em IEEE Transactions on Industrial Informatics}, 17(4):2821--2832,
  2020.

\bibitem{wang2020deep}
Jingdong Wang, Ke Sun, Tianheng Cheng, Borui Jiang, Chaorui Deng, Yang Zhao,
  Dong Liu, Yadong Mu, Mingkui Tan, Xinggang Wang, et~al.
\newblock Deep high-resolution representation learning for visual recognition.
\newblock {\em IEEE transactions on pattern analysis and machine intelligence},
  2020.

\bibitem{wofk2019fastdepth}
Diana Wofk, Fangchang Ma, Tien-Ju Yang, Sertac Karaman, and Vivienne Sze.
\newblock Fastdepth: Fast monocular depth estimation on embedded systems.
\newblock In {\em 2019 International Conference on Robotics and Automation
  (ICRA)}, pages 6101--6108. IEEE, 2019.

\bibitem{wu2019fbnet}
Bichen Wu, Xiaoliang Dai, Peizhao Zhang, Yanghan Wang, Fei Sun, Yiming Wu,
  Yuandong Tian, Peter Vajda, Yangqing Jia, and Kurt Keutzer.
\newblock Fbnet: Hardware-aware efficient convnet design via differentiable
  neural architecture search.
\newblock In {\em Proceedings of the IEEE conference on computer vision and
  pattern recognition}, pages 10734--10742, 2019.

\bibitem{yang2021transformers}
Guanglei Yang, Hao Tang, Mingli Ding, Nicu Sebe, and Elisa Ricci.
\newblock Transformers solve the limited receptive field for monocular depth
  prediction.
\newblock {\em arXiv preprint arXiv:2103.12091}, 2021.

\bibitem{yang2018netadapt}
Tien-Ju Yang, Andrew Howard, Bo Chen, Xiao Zhang, Alec Go, Mark Sandler,
  Vivienne Sze, and Hartwig Adam.
\newblock Netadapt: Platform-aware neural network adaptation for mobile
  applications.
\newblock In {\em Proceedings of the European Conference on Computer Vision
  (ECCV)}, pages 285--300, 2018.

\bibitem{yang2019lightweight}
Wenming Yang, Wei Wang, Xuechen Zhang, Shuifa Sun, and Qingmin Liao.
\newblock Lightweight feature fusion network for single image super-resolution.
\newblock {\em IEEE Signal Processing Letters}, 26(4):538--542, 2019.

\bibitem{Yin2019enforcing}
Wei Yin, Yifan Liu, Chunhua Shen, and Youliang Yan.
\newblock Enforcing geometric constraints of virtual normal for depth
  prediction.
\newblock In {\em The IEEE International Conference on Computer Vision (ICCV)},
  2019.

\bibitem{yu2021bisenet}
Changqian Yu, Changxin Gao, Jingbo Wang, Gang Yu, Chunhua Shen, and Nong Sang.
\newblock Bisenet v2: Bilateral network with guided aggregation for real-time
  semantic segmentation.
\newblock {\em International Journal of Computer Vision}, 129(11):3051--3068,
  2021.

\bibitem{yu2018bisenet}
Changqian Yu, Jingbo Wang, Chao Peng, Changxin Gao, Gang Yu, and Nong Sang.
\newblock Bisenet: Bilateral segmentation network for real-time semantic
  segmentation.
\newblock In {\em Proceedings of the European conference on computer vision
  (ECCV)}, pages 325--341, 2018.

\bibitem{yu2021lite}
Changqian Yu, Bin Xiao, Changxin Gao, Lu Yuan, Lei Zhang, Nong Sang, and
  Jingdong Wang.
\newblock Lite-hrnet: A lightweight high-resolution network.
\newblock In {\em Proceedings of the IEEE conference on computer vision and
  pattern recognition}, pages 10440--10450, 2021.

\bibitem{yucel2021real}
Mehmet~Kerim Yucel, Valia Dimaridou, Anastasios Drosou, and Albert Saa-Garriga.
\newblock Real-time monocular depth estimation with sparse supervision on
  mobile.
\newblock In {\em Proceedings of the IEEE conference on computer vision and
  pattern recognition}, pages 2428--2437, 2021.

\bibitem{zeyde2010single}
Roman Zeyde, Michael Elad, and Matan Protter.
\newblock On single image scale-up using sparse-representations.
\newblock In {\em International conference on curves and surfaces}, pages
  711--730. Springer, 2010.

\bibitem{zhang2018image}
Yulun Zhang, Kunpeng Li, Kai Li, Lichen Wang, Bineng Zhong, and Yun Fu.
\newblock Image super-resolution using very deep residual channel attention
  networks.
\newblock In {\em Proceedings of the European conference on computer vision
  (ECCV)}, pages 286--301, 2018.

\bibitem{zhao2018icnet}
Hengshuang Zhao, Xiaojuan Qi, Xiaoyong Shen, Jianping Shi, and Jiaya Jia.
\newblock Icnet for real-time semantic segmentation on high-resolution images.
\newblock In {\em Proceedings of the European conference on computer vision
  (ECCV)}, pages 405--420, 2018.

\bibitem{zhao2017pyramid}
Hengshuang Zhao, Jianping Shi, Xiaojuan Qi, Xiaogang Wang, and Jiaya Jia.
\newblock Pyramid scene parsing network.
\newblock In {\em Proceedings of the IEEE conference on computer vision and
  pattern recognition}, pages 2881--2890, 2017.

\bibitem{zheng2015conditional}
Shuai Zheng, Sadeep Jayasumana, Bernardino Romera-Paredes, Vibhav Vineet,
  Zhizhong Su, Dalong Du, Chang Huang, and Philip~HS Torr.
\newblock Conditional random fields as recurrent neural networks.
\newblock In {\em Proceedings of the IEEE international conference on computer
  vision}, pages 1529--1537, 2015.

\bibitem{zhou2020rethinking}
Daquan Zhou, Qibin Hou, Yunpeng Chen, Jiashi Feng, and Shuicheng Yan.
\newblock Rethinking bottleneck structure for efficient mobile network design.
\newblock In {\em Computer Vision--ECCV 2020: 16th European Conference,
  Glasgow, UK, August 23--28, 2020, Proceedings, Part III 16}, pages 680--697.
  Springer, 2020.

\bibitem{zhu2019efficient}
Feiyang Zhu and Qijun Zhao.
\newblock Efficient single image super-resolution via hybrid residual feature
  learning with compact back-projection network.
\newblock In {\em Proceedings of the IEEE/CVF International Conference on
  Computer Vision Workshops}, pages 0--0, 2019.

\bibitem{zoph2016neural}
Barret Zoph and Quoc~V Le.
\newblock Neural architecture search with reinforcement learning.
\newblock {\em arXiv preprint arXiv:1611.01578}, 2016.

\bibitem{zoph2018learning}
Barret Zoph, Vijay Vasudevan, Jonathon Shlens, and Quoc~V Le.
\newblock Learning transferable architectures for scalable image recognition.
\newblock In {\em Proceedings of the IEEE conference on computer vision and
  pattern recognition}, pages 8697--8710, 2018.

\end{thebibliography}
}

\end{document}